\documentclass[sigconf]{acmart}

\AtBeginDocument{%
  }

\copyrightyear{2022} 
\acmYear{2022} 
\setcopyright{acmlicensed}\acmConference[MMAsia '22]{ACM Multimedia Asia}{December 13--16, 2022}{Tokyo, Japan}
\acmBooktitle{ACM Multimedia Asia (MMAsia '22), December 13--16, 2022, Tokyo, Japan}
\acmPrice{15.00}
\acmDOI{10.1145/3551626.3564941}
\acmISBN{978-1-4503-9478-9/22/12}

\acmSubmissionID{36}

\usepackage[labelformat=simple]{subcaption}
\newcommand{\R}{\mathbb{R}}
\newcommand{\tabscaleA}{0.9}
\newcommand{\tabscaleB}{0.8}

\begin{document}

\title[ObjectMix: Data Augmentation by Copy-Pasting Objects in Videos for Action Recognition]%
{ObjectMix: Data Augmentation by Copy-Pasting\\ Objects in Videos for Action Recognition}




\author{Jun Kimata}
\affiliation{%
 \institution{Nagoya Institute of Technology}
 \country{Japan}}

\author{Tomoya Nitta}
\affiliation{%
 \institution{Nagoya Institute of Technology}
 \country{Japan}}

\author{Toru Tamaki}
\affiliation{%
 \institution{Nagoya Institute of Technology}
 \country{Japan}}



\renewcommand{\shortauthors}{Kimata et al.}

\begin{abstract}
In this paper, we propose a data augmentation method for action recognition using instance segmentation.
Although many data augmentation methods have been proposed for image recognition, few of them are tailored for action recognition.
Our proposed method, ObjectMix, extracts each object region from two videos using instance segmentation
and combines them to create new videos.
Experiments on two action recognition datasets, UCF101 and HMDB51, demonstrate the effectiveness of the proposed method
and show its superiority over VideoMix, a prior work.

\end{abstract}



\maketitle

\section{Introduction}

In recent years, there has been a lot of research on video recognition, which are used in various applications.
One of the problems in developing action recognition models is the cost of constructing datasets
\cite{DBLP:ucf101,DBLP:journals/corr/Kinetics,wishart2018hmdb}. 
In actual application scenarios, 
practitioners often need to prepare a new dataset for their tasks, but
the annotation cost of labeling a large number of videos is inherently high,
and for some applications it may not be possible to collect many videos in the first place.
In which cases, training on a small dataset is inevitable.

There are three approaches to the small dataset issue.
The first is to synthesize various images with 3D models,
for example, pose estimation \cite{varol17_surreal}
and flow estimation \cite{Sintel,FlyingChairs,Flyingthings3D}.
This approach has the advantage of being able to generate as many images as possible,
while it can only be applied to tasks where images can be easily synthesized.
For action recognition, it is necessary to consider various objects in scenes, motions and actions of people,
but currently no methods are available to generate such realistic action videos.

The second is to use GAN \cite{GAN,9528943} to generate images.
Once a GAN model has been trained,
it can generate any number of realistic images.
This approach has been used as data generation in medical image processing 
where it is not possible to collect data at a large scale \cite{cGAN_Medical,SA-GAN_CT}.
GANs are good at generating images with specific structures 
such as faces, human bodies, animals, and indoor images.
But it is still difficult to generate videos of various scenes \cite{GAN_video,MocoGAN}
such as those appear in videos for action recognition.

The third is data augmentation
\cite{DBLP:journals/air/KhalifaLM22,DBLP:journals/jbd/ShortenK19,DBLP:journals/corr/abs-2205-01491},
which is widely used because of its simplicity.
Data augmentation refers to applying various image processing to images
such as flipping, rotation, contrast change, and adding various noises \cite{imgaug,Albumentations}.
This means that even the dataset size is small,
the generalization performance of the model is expected to be as good as when trained on a large dataset.
Recently, mix-type methods \cite{Zhang_iclr18,DBLP:journals/corr/abs-1708-04552,Yun_2019_ICCV} have been proposed
that apply operations such as cropping a portion of an image, 
pasting the portion onto another image, and blending these two images and their corresponding labels.
In addition, task-specific methods have also been proposed, 
such as for monocular depth estimation \cite{CutDepth},
super-resolution \cite{CutBlur},
object detection \cite{cutpaste},
and instance segmentation \cite{CopyPaste}.
However, few are specific to action recognition \cite{Video_Action_Understanding2021}.
Commonly used data augmentation for action recognition is simply applying the same geometric and photometric augmentation to all frames at once,
except vertical flipping because usually videos are not recorded upside down.
Also, for some datasets horizontal flipping is also not used since some actions distinguish left and right;
for example, in something-something v2 (SSv2) \cite{SSv2}, ``move from right to left'' and ``move from left to right'' are different categories.
An exception is VideoMix \cite{VideoMix}, a recently proposed mix-type data augmentation method specialized for action recognition.
However, this is a simple application of CutMix \cite{Yun_2019_ICCV} to the spatio-temporal 3D volume of video frames,
and does not take into any consideration the temporal and spatial continuity of the video contents.

In this study, we propose a method that extends Copy-Paste \cite{CopyPaste} to action recognition.
Copy-Paste is a mix-type method
that generates new images by extracting object regions from two images by semantic segmentation,
and then pasting the objects onto each other using the other image as a background.
The proposed method performs it to videos, in other words, object regions from each of the two video frames and paste them into each other's video frames to create new ones.
The contributions of this paper are as follows.
\begin{itemize}

\item We propose a new mix-type method of data augmentation for action recognition.
It is an extension of Copy-Paste, and unlike VideoMix, it is possible to generate video frames that take into consideration the temporal and spatial continuity of objects in the videos.

\item The proposed method is applicable to any existing action recognition models.
This allows the recognition performance of existing models
to be improved using the proposed method.

\item In experiments using two action recognition datasets,
we show that the performance of the proposed method 
is better than that of VideoMix.

\end{itemize}

\begin{figure}[t]
    \centering

    \begin{minipage}[t]{0.9\linewidth}
    \centering
    \includegraphics[width=\linewidth]{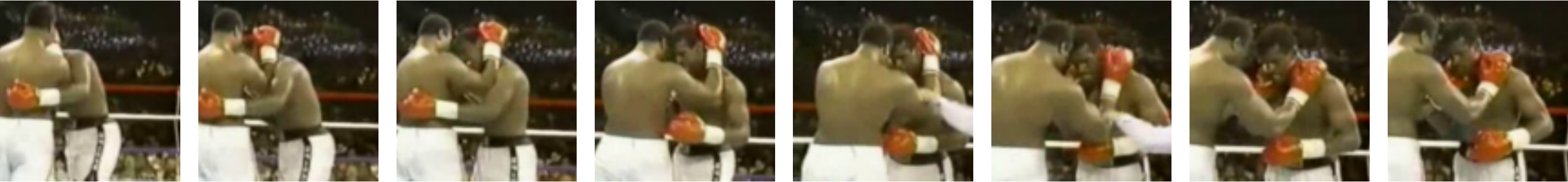}
    \subcaption{}
    \label{fig:ovmixbase1}
    \end{minipage}
    \vspace{0.5em}

    \begin{minipage}[t]{0.9\linewidth}
    \centering
    \includegraphics[width=\linewidth]{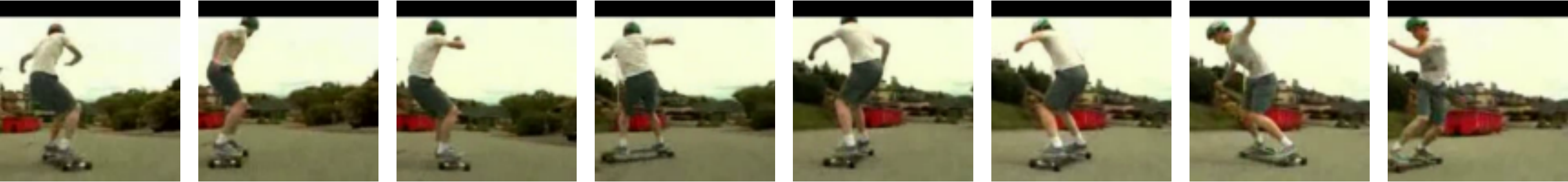}
    \subcaption{}
    \label{fig:ovmixbase2}
    \end{minipage}
    \vspace{0.5em}
    
    \begin{minipage}[t]{0.9\linewidth}
    \centering
    \includegraphics[width=\linewidth]{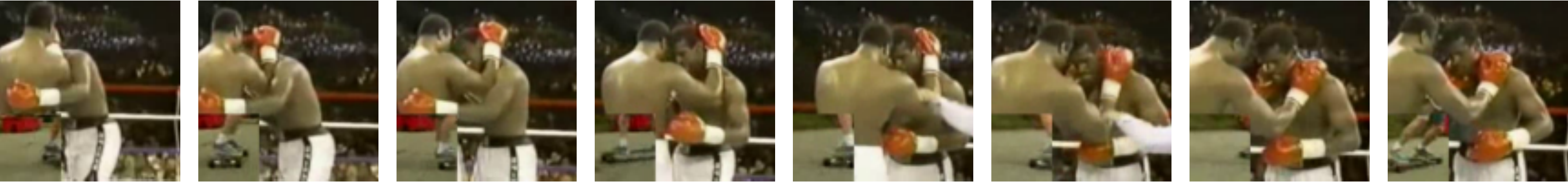}
    \subcaption{}
    \label{fig:ovmix1}
    \end{minipage}
    \vspace{0.5em}

    \begin{minipage}[t]{0.9\linewidth}
    \centering
    \includegraphics[width=\linewidth]{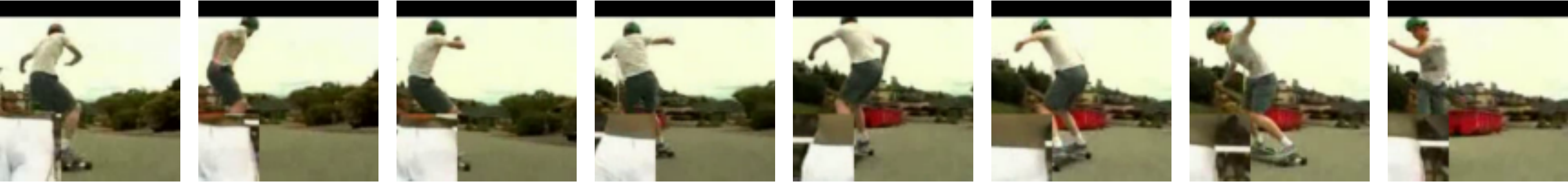}
    \subcaption{}
    \label{fig:ovmix2}
    \end{minipage}

    \caption{
    Example of video generation with VideoMix.
    (a)(b) Two original videos and
    (c)(d) two generated Videos.
    }

    \label{fig:VideoMix}
\end{figure}

\begin{figure}[t]
    \centering

    \begin{minipage}[t]{0.9\linewidth}
    \centering
    \includegraphics[width=\linewidth]{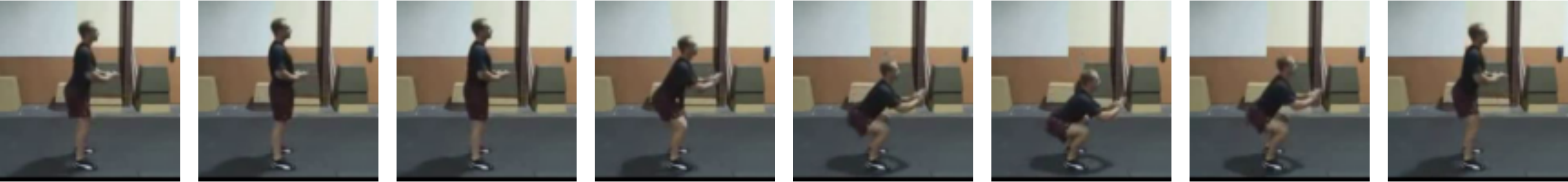}
    \subcaption{}
    \label{fig:base1}
    \end{minipage}
    \vspace{0.5em}

    \begin{minipage}[t]{0.9\linewidth}
    \centering
    \includegraphics[width=\linewidth]{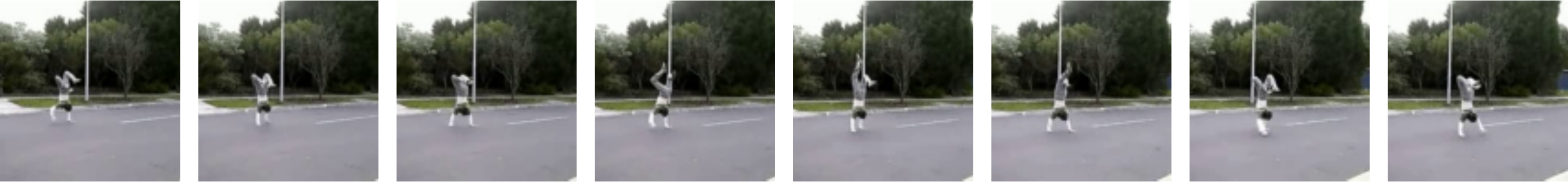}
    \subcaption{}
    \label{fig:base2}
    \end{minipage}
    \vspace{0.5em}
    
    \begin{minipage}[t]{0.9\linewidth}
    \centering
    \includegraphics[width=\linewidth]{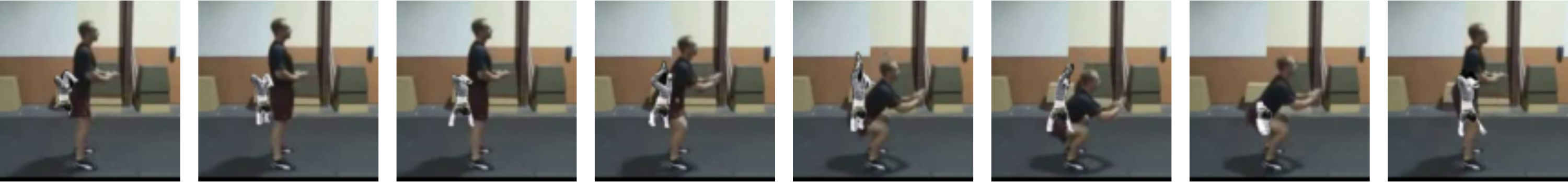}
    \subcaption{}
    \label{fig:mix1}
    \end{minipage}
    \vspace{0.5em}

    \begin{minipage}[t]{0.9\linewidth}
    \centering
    \includegraphics[width=\linewidth]{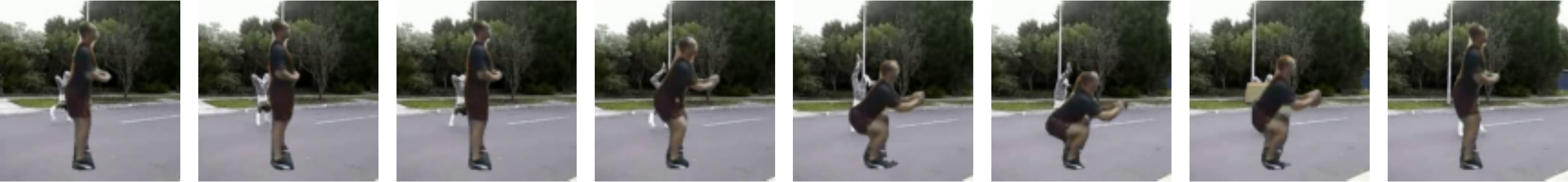}
    \subcaption{}
    \label{fig:mix2}
    \end{minipage}

    \caption{
    Example of video generation with ObjectMix.
    (a)(b) Two original videos and
    (c)(d) two generated Videos.
    }

    \label{fig:ObjectMix}
\end{figure}

\section{Related Works}

\noindent\textbf{Data augmentation}
has been widely used to improve the performance
by augmenting training samples with various transformations
\cite{DBLP:journals/air/KhalifaLM22,DBLP:journals/jbd/ShortenK19,DBLP:journals/corr/abs-2205-01491}.
In addition to simple image processing such as rotation, translation, noise \cite{imgaug,Albumentations},
there are also a number of mix-type methods
(Mixup \cite{Zhang_iclr18}, 
Cutout \cite{DBLP:journals/corr/abs-1708-04552}, 
CutMix \cite{Yun_2019_ICCV}),
and task-specific methods
(CutDepth \cite{CutDepth}, 
CutBlur \cite{CutBlur},
Cut-Paste-Learn \cite{cutpaste}).
Copy-Paste \cite{CopyPaste} is a simple method for the task of instance segmentation:
for two images, it cuts out only the instance region of in one image, and randomly pastes it onto the other image.
Advantages of this method include the fact that spatial continuity is guaranteed
since the whole instances are always pasted, and that further augmentation, such as scaling, can be performed on the pasted instances.

\noindent\textbf{Action recognition} 
is the task of identifying human actions in a video \cite{Video_Action_Understanding2021}.
Unlike image recognition, action recognition requires to model temporal information.
For example, Two-Stream types \cite{TwoStream} use optical flow as input as the temporal information,
and 3D CNN methods such as
3D ResNet \cite{3Dresnet}, X3D \cite{X3D}, and SlowFast \cite{SlowFast}
perform 3D convolution in spatio-temporal volumes.
Usually, models pre-trained on large datasets (Kinetics \cite{DBLP:journals/corr/Kinetics})
are transferred to small datasets (UCF101 \cite{DBLP:ucf101} and HMDB51 \cite{wishart2018hmdb}).
However, even when fine-tuning on small datasets, the training set should be diverse,
and in such cases,
data augmentation would also be important to ensure better generalization performance.

\noindent\textbf{VideoMix.}
The above data augmentation methods were proposed for image recognition tasks.
Few methods exist for action recognition, with the sole exception of VideoMix \cite{VideoMix},
whose example is shown in Figure \ref{fig:VideoMix}.
This method is a direct extension of CutMix \cite{Yun_2019_ICCV} to 3D video volumes,
whereby a cube is cut from one video volume and pasted to another.
The problem here is the discontinuity of objects in the video.
Since the location of the cube is randomly selected,
the entire regions of objects and humans may not be shown in the pasted video,
or even worse, no objects might appear in the cube.
Furthermore, the objects that appear in the beginning may disappear in the middle of the video.
In contrast, as shown in Figure \ref{fig:ObjectMix},
the proposed method does not break the continuity of the objects and humans to be pasted,
and keeps the objects shown for all the frames.

\section{Method}

This section describes an overview of the proposed method that
consists of the following processes.
\begin{enumerate}
    \item Preparing two source videos $v_1$ and $v_2$. Let $y_1, y_2$ be the labels of each.
    \item Extracting object regions from each video and generating object masks $M_1$ and $M_2$.
    \item Pasting the masked region of one video onto the other video to create new videos $v_{12}$ and $v_{21}$.
    \item Using the mask information to generate labels $y_{12}$ and $y_{21}$.
\end{enumerate}
Examples of videos generated by the proposed method are shown in Figures \ref{fig:ObjectMix} (c) and (d).

\subsection{Video preparation}

The source videos $v_1, v_2 \in \R^{T \times C \times H \times W}$ are video clips consisting of
$T$ frames $v_1(t), v_2(t) \in \R^{C \times H \times W}$ for $t=1,\ldots,T$,
where $H, W$ are height and width of the frames.
Let $y_1, y_2 \in \{0, 1\}^{L_a}$ be the corresponding one-hot encoded labels,
where $L_a$ is the number of categories.

\subsection{Extracting object masks}

For each frame $v_k(t)$ for $k=1,2$, we apply instance segmentation 
to generate masks $M_k(t) \in \{0, 1\}^{N_k(t) \times H \times W}$ for $t=1,\ldots,T$
where $N_k(t)$ is the number of instances detected in $v_k(t)$.

\begin{figure}[t]
    \centering
    
    \begin{minipage}[t]{0.9\linewidth}
    \centering
    \includegraphics[width=\linewidth]{images/ims/ori0_1.pdf}
    \subcaption{}
    \label{fig:mask_base}
    \end{minipage}
    \vspace{0.5em}

    \begin{minipage}[t]{0.9\linewidth}
    \centering
    \includegraphics[width=\linewidth]{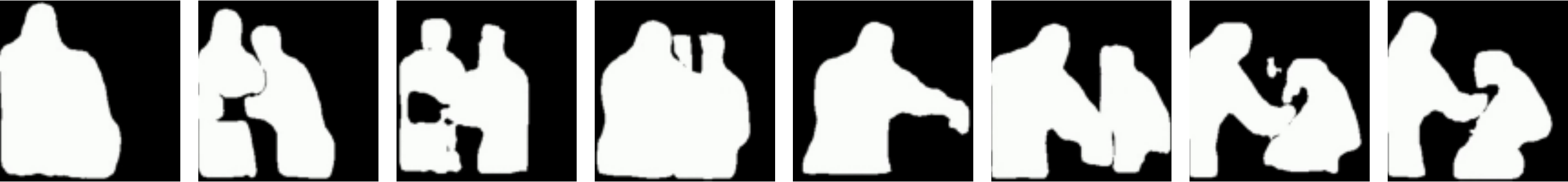}
    \subcaption{}
    \label{fig:mask}
    \end{minipage}
    \vspace{0.5em}
    
    \begin{minipage}[t]{0.9\linewidth}
    \centering
    \includegraphics[width=\linewidth]{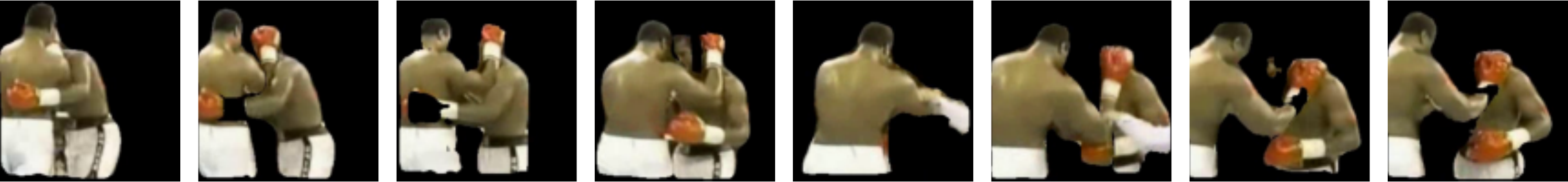}
    \subcaption{}
    \label{fig:ins}
    \end{minipage}
    \vspace{0.5em}
    
    \begin{minipage}[t]{00.9\linewidth}
    \centering
    \includegraphics[width=\linewidth]{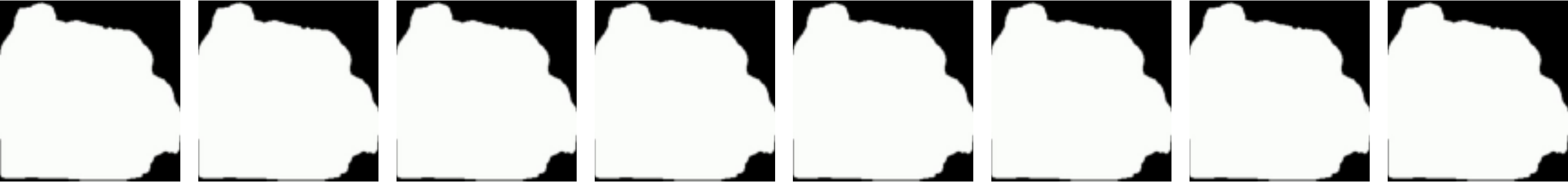}
    \subcaption{}
    \label{fig:mask_or}
    \end{minipage}
    \vspace{0.5em}
    
    \begin{minipage}[t]{0.9\linewidth}
    \centering
    \includegraphics[width=\linewidth]{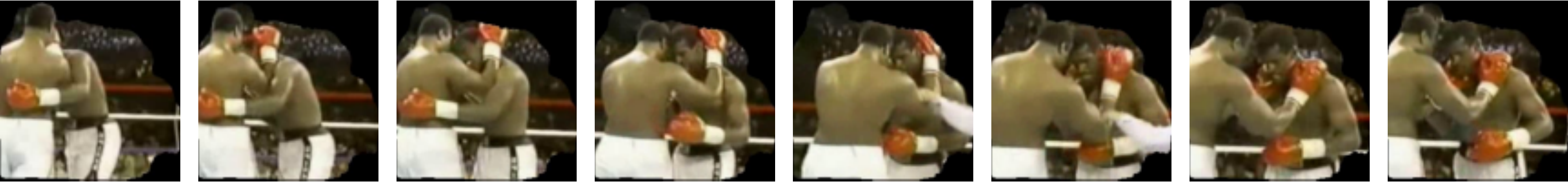}
    \subcaption{}
    \label{fig:ins_or}
    \end{minipage}

    \caption{
    Example of extracted object masks.
    (a) Original video.
    (b) Mask $M_1'$ and
    (c) corresponding objects.
    (d) Mask $M_1''$ and
    (e) corresponding objects.
    }

    \label{fig:mask_example}
\end{figure}

If multiple instances are extracted (i.e., $N_k(t) > 1$),
they are aggregated into a single-channel mask $M'_k(t) \in \{0, 1\}^{1 \times H \times W}$
by logical OR as follows;
\begin{align}
    M'_k(t) &= \bigcup_{n=1}^{N_k(t)} M_{k,n}(t),
\end{align}
where $M_{k,n}(t)$ is the $n$-th channel of $M_k(t)$.
Examples of this mask and extracted objects are shown in Figure \ref{fig:mask}.
We used Detectron2 \cite{wu2019detectron2} that are 
pre-trained on the COCO \cite{MSCOCO} dataset with 80 classes
(therefore $0 \le N(t) \le 80$).
In this study, all extracted instances are used for mask generation,
regardless of the relevance of the extracted instance categories to the action categories.

\subsection{Temporal aggregation of masks}

Masks are extracted from each frame, however,
the temporal continuity of the masks would be lost
if the instance segmentation fails to detect objects in a certain frame
as shown in fifth column of Figure \ref{fig:mask}.
Therefore,
we propose to aggregate the masks of each extracted frame by logical OR in the temporal direction as well.
\begin{align}
    M''_k(t) &= \bigcup_{t=1}^{T} \bigcup_{n=1}^{N_k(t)} M_{k,n}(t).
\end{align}
The masks $M''_k$ are the same for all frames,
however, even in frames where detection fails, the mask of the object can still be extracted.
An example of this mask $M_k''$ is shown in Figure \ref{fig:mask_or}.
In our experiments, we refer to the proposed method with $M_k'$ as ObjectMix,
and the version with $M_k''$ as ObjectMix+or.

\subsection{Video and Label Composition}

Next, we generate new videos using the generated masks $M_1', M_2'$ (or $M_1'', M_2''$).
In this case, two videos can be generated,
that is, the object extracted with mask $M_1'$ from one video $v_1$ is pasted onto the other video $v_2'$, and vice versa.
\begin{align}
    v_{12}(t) &= v_1'(t) \odot M_1'(t) + v_2'(t) \odot (1 - M_1'(t)) \\
    v_{21}(t) &= v_1'(t) \odot (1 - M_2'(t)) + v_2'(t) \odot M_2'(t),
\end{align}
where $\odot$ is the element-wise product.

To define weights for label composition, we use the fraction of pixels with non-zero values
in the generated object masks.
First, we define weights as
$\lambda_k = \frac{|M_k'|}{THW}$,
where $|M'_k|$ is the sum of non-zero values in the binary mask $M'_k$.
Then, we composite the labels as follows;
\begin{align}
    y_{12} &= \lambda_1 y_1 + (1 - \lambda_1) y_2 \\
    y_{21} &= (1 - \lambda_2) y_1 + \lambda_2 y_2.
\end{align}

This is similar to CutMix \cite{Yun_2019_ICCV},
however the weights of CutMix are fixed in advance,
and the rectangle whose ratio of the rectangle's area to the entire image
matches the weight is randomly selected.
In contrast, the weights of the proposed method are dynamically adjusted according to the area of objects in the mask.

\subsection{Loss}

To train a model,
we compute the cross-entropy (CE) loss.
Let model predictions be $\hat{y}_{ij}, \hat{y}_{ji}$ for videos $v_{ij}, v_{ji}$
generated from source videos $v_i, v_j$.
The CE losses
$L_\mathrm{CE} (\hat{y}_{ij}, y_{ij})$
or
$L_\mathrm{CE} (\hat{y}_{ji}, y_{ji})$
need to be computed with an appropriate weight.

A typical implementation of mix-type augmentation
computes the loss on a batch basis.
Assuming that videos $v_i, v_j$ are in the batch of size $B$,
the loss for the batch is calculated as follows;
\begin{align}
    \lambda \sum_{i=1}^B L_\mathrm{CE} (\hat{y}_{ij_i}, y_{i}) +
    (1 - \lambda) \sum_{i=1}^B L_\mathrm{CE} (\hat{y}_{ij_i}, y_{j_i}),
\end{align}
where $j_1,\ldots,j_B$ is a certain permutation of $1,\ldots,B$.

However, in the case of the proposed method,
and the weights are the relative area of the masks and
different for each sample in the batch.
Therefore, we compute the following loss;
\begin{align}
    \sum_{i=1}^B
    \lambda_{i} L_\mathrm{CE} (\hat{y}_{ij_i}, y_{i}) +
    (1 - \lambda_{i}) L_\mathrm{CE} (\hat{y}_{ij_i}, y_{j_i}).
\end{align}

\section{experimental results}

In this section
we report experimental results with two action recognition datasets
to evaluate the performance of the proposed method and compare it with VideoMix.

\subsection{Datasets}

The following two datasets were used.

UCF101 \cite{DBLP:ucf101}
has 101 classes of human actions, consisting of a training set of about 9500 videos and a validation set of about 3500 videos.
Each video was collected from Youtube, 
with an average length of 7.21 seconds.
There are three splits for training and validation, 
and we report the performance of the first split as it is usually used.

HMDB51 \cite{wishart2018hmdb}
has 51 classes of human actions,
consisting of a training set of 3570 videos and a validation set of 1530 videos.
Each video is collected from movies, Web, Youtube, etc.,
and the average length is 3.15 seconds.
There are three splits for training and validation, and we used the first split.

\subsection{Experimental Settings}

We used X3D-M \cite{X3D}, a 3D CNN-based action recognition model,
pre-trained on Kinetics400 \cite{DBLP:journals/corr/Kinetics}.
In training, we randomly sampled 16 frames per clip from the video,
and randomly determined the short side of the frame in the range of [224, 320] pixels and
resized it while maintaining the aspect ratio,
and randomly cropped a $224\times224$ pixel patch,
then flipped horizontally with a probability of 50\%.
No photometric augmentation were used.
The optimizer was Adam with the learning rate of 0.0001 and the batch size of 16.
Training epochs were set to 10 for UCF101 and 20 for HMDB51,
so that the top-1 performance for the training set would roughly converge.

We used a single view test for validation (i.e., one clip was randomly sampled from a single video) instead of the multi-view test \cite{Nonlocal}.
Frames were resized so that the short side of the frame is 256 pixels
while maintaining the aspect ratio, and the central $224\times224$ pixels of the frame were cropped.
To take into account the randomness of the clip sampling,
we report the mean and standard deviation of 10 results.

Augmentation was randomly applied to each batch with the probability $0 \le p \le 1$.
Note that $p=0$ is equivalent to the case where no augmentation is applied.
In the following experiments, performances are reported for $p=0, 0.2, \ldots, 1$. 0.2.

\subsection{Results of ObjectMix}

\begin{table}[t]

\centering

\caption{The top-1 performance of ObjectMix (OM) on UCF101 and HMDB51 validation set.
The $p=0.0$ is the baseline without applying the proposed method.
}
\label{tab:ucf_aug}
\label{tab:hmdb_aug}

\scalebox{\tabscaleA}{
\begin{tabular}{c|cc|cc}
\multicolumn{1}{c|}{} & \multicolumn{2}{c|}{UCF101} & \multicolumn{2}{c}{HMDB51} \\
method (p)      & top-1 & top-5 & top-1 & top-5 \\ \hline
OM (0.0)  & $93.58 \pm  0.03$ & $99.23 \pm 0.01$ & $69.86 \pm  0.39$ & $91.89 \pm 0.26$ \\ \hline
OM (0.2)  & $94.68 \pm  0.23$ & $\textbf{99.65} \pm 0.05$ & $70.59 \pm  0.37$ & $93.18 \pm 0.54$ \\
OM (0.4)  & $93.63 \pm  0.24$ & $99.40 \pm 0.08$ & $\textbf{71.47} \pm  0.30$ & $92.77 \pm 0.31$ \\
OM (0.6)  & $\textbf{95.12} \pm  0.15$ & $99.46 \pm 0.07$ & $71.44 \pm  0.59$ & $\textbf{93.66} \pm 0.28$ \\
OM (0.8)  & $93.94 \pm  0.23$ & $99.22 \pm 0.05$ & $70.74 \pm  0.29$ & $92.38 \pm 0.26$ \\
OM (1.0)  & $93.45 \pm  0.21$ & $98.96 \pm 0.07$ & $69.43 \pm  0.46$ & $92.49 \pm 0.28$ \\
\end{tabular}
}

\end{table}

First, we show the comparison of ObjectMix ($p>0$) with no augmentation ($p=0$)
in Table \ref{tab:ucf_aug}.
As $p$ increases, the performance tends to have a peak at $p=0.6$ or $0.4$,
then deteriorates for larger values.
This is probably due to the fact that
the original videos are used less for training for large values of $p$.
In other words, the original and generated videos should be balanced.

\begin{figure}[t]
    \centering

    \begin{minipage}[t]{0.49\linewidth}
    \centering
    \includegraphics[width=\linewidth]{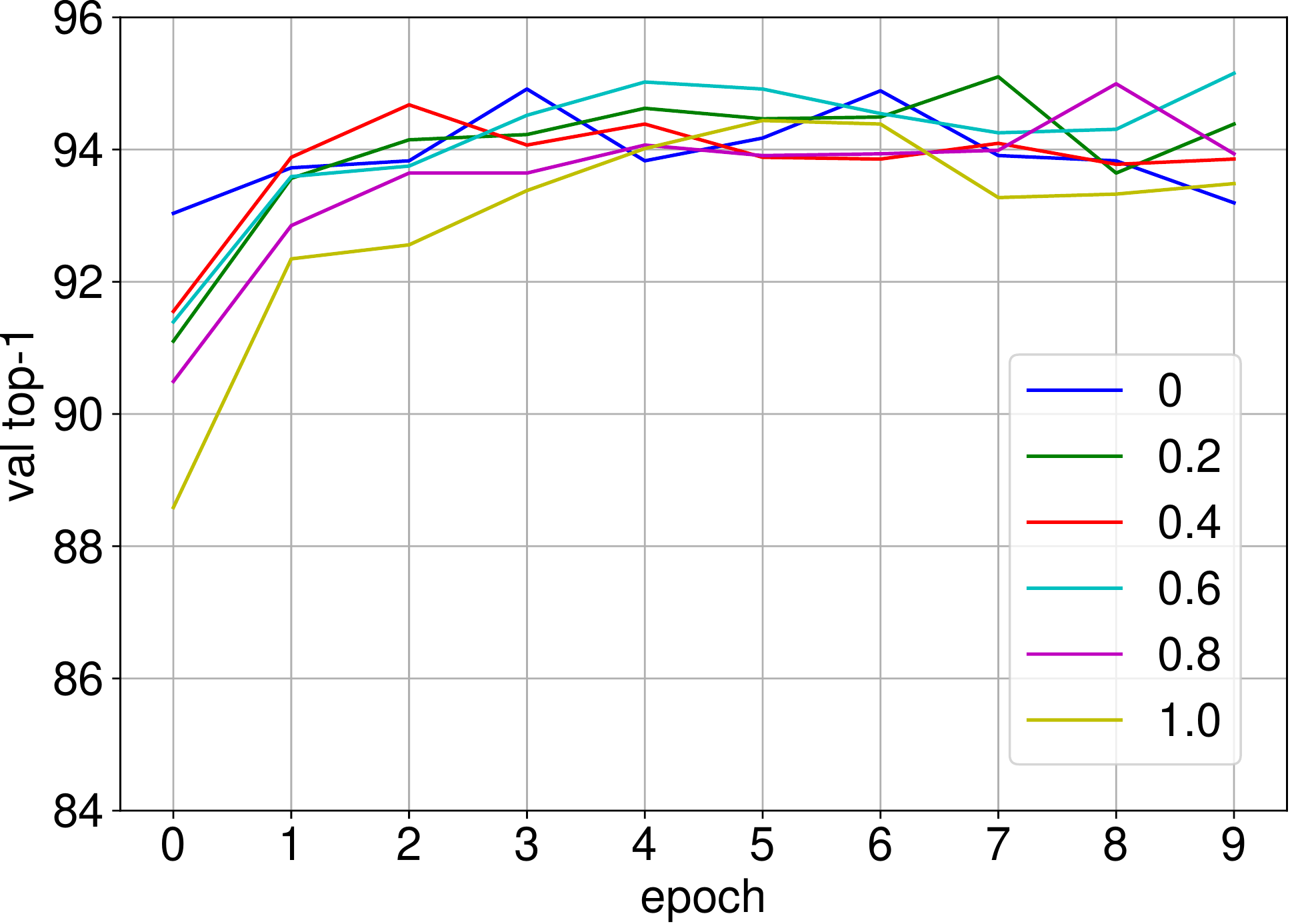}
    \end{minipage}
    \hfill
    \begin{minipage}[t]{0.49\linewidth}
    \centering
    \includegraphics[width=\linewidth]{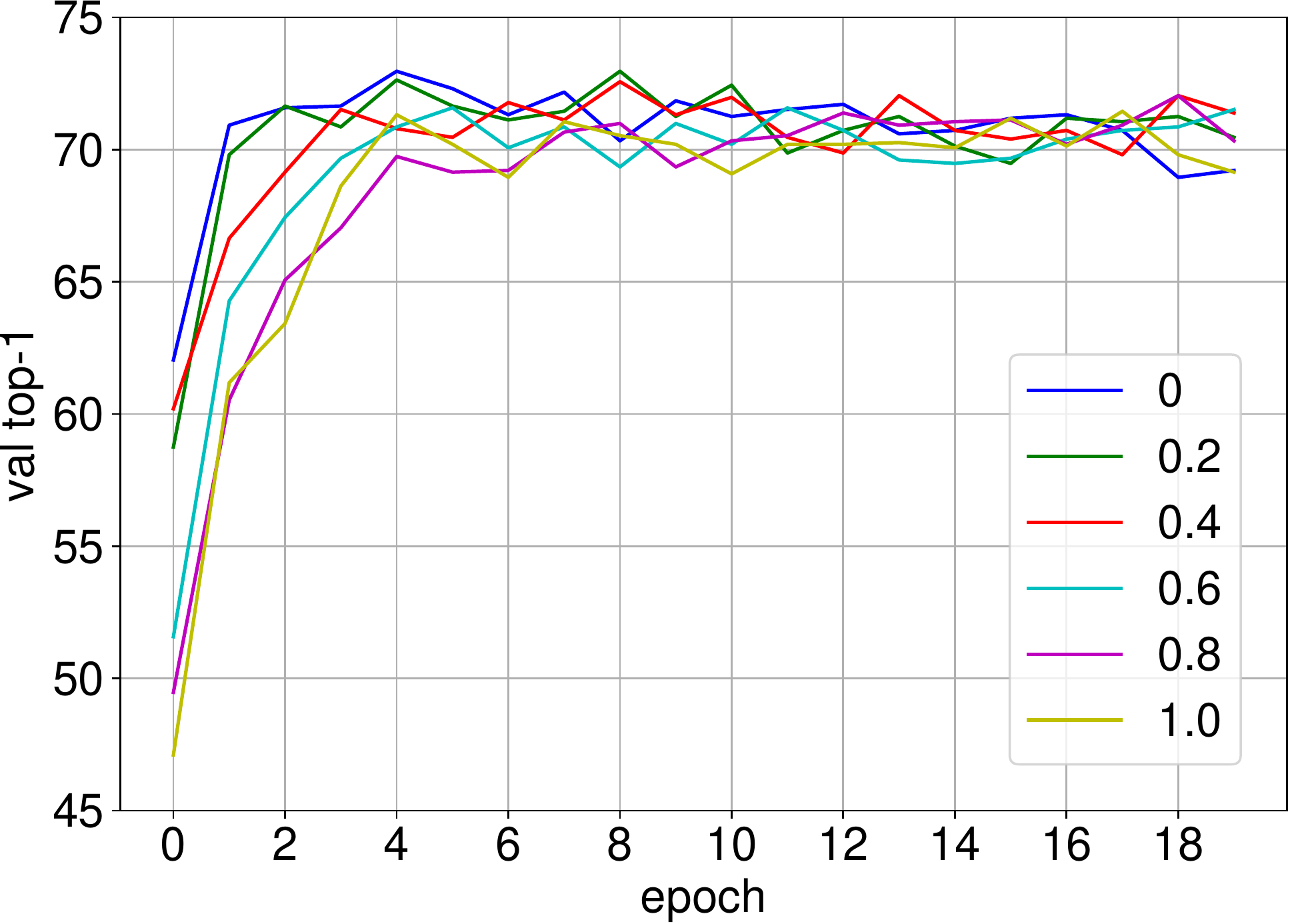}
    \end{minipage}

    \begin{minipage}[t]{0.49\linewidth}
    \centering
    \includegraphics[width=\linewidth]{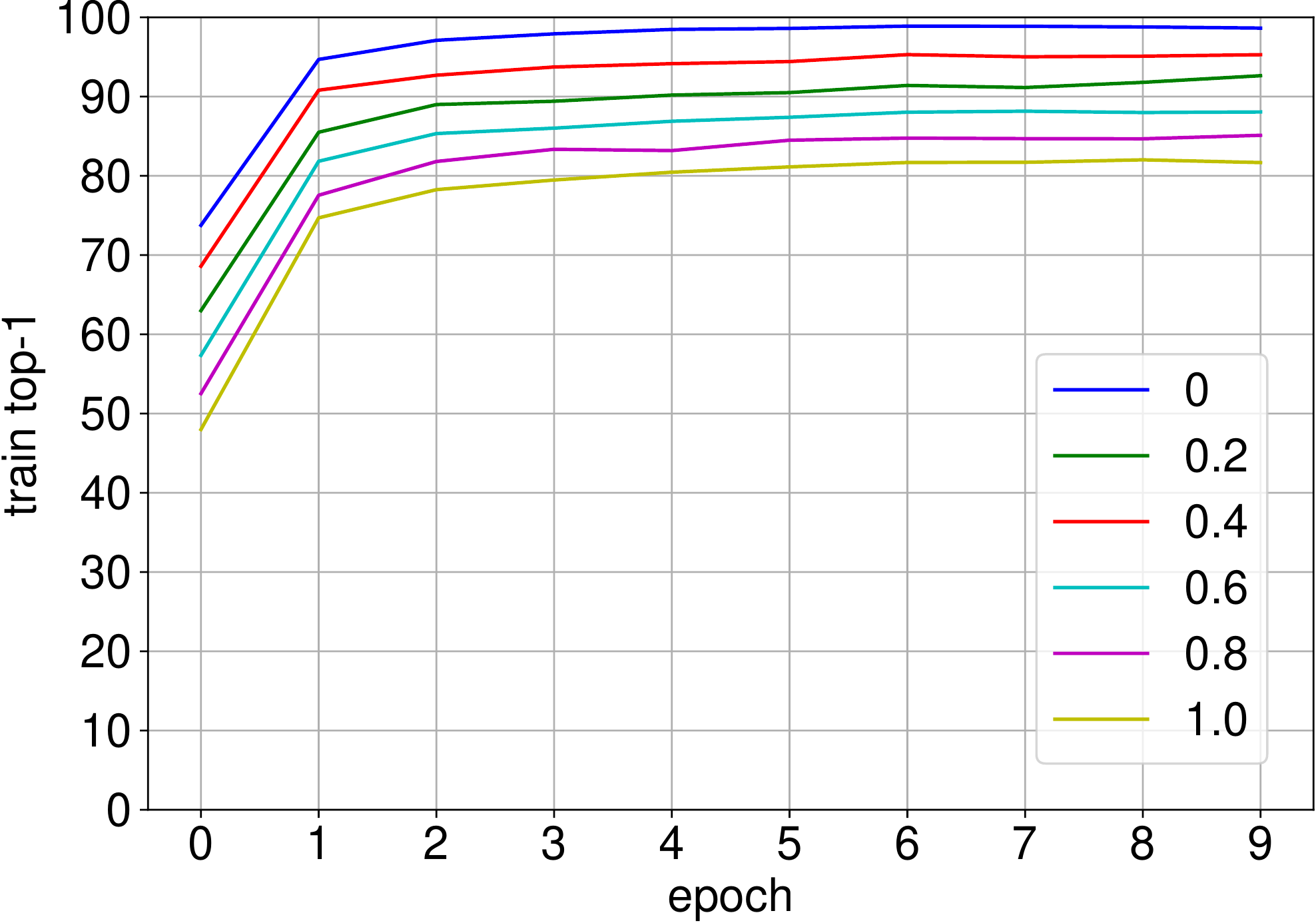}
    \subcaption{UCF101}
    \label{fig:hmdb_aug_acc_t}
    \end{minipage}
    \hfill
    \begin{minipage}[t]{0.49\linewidth}
    \centering
    \includegraphics[width=\linewidth]{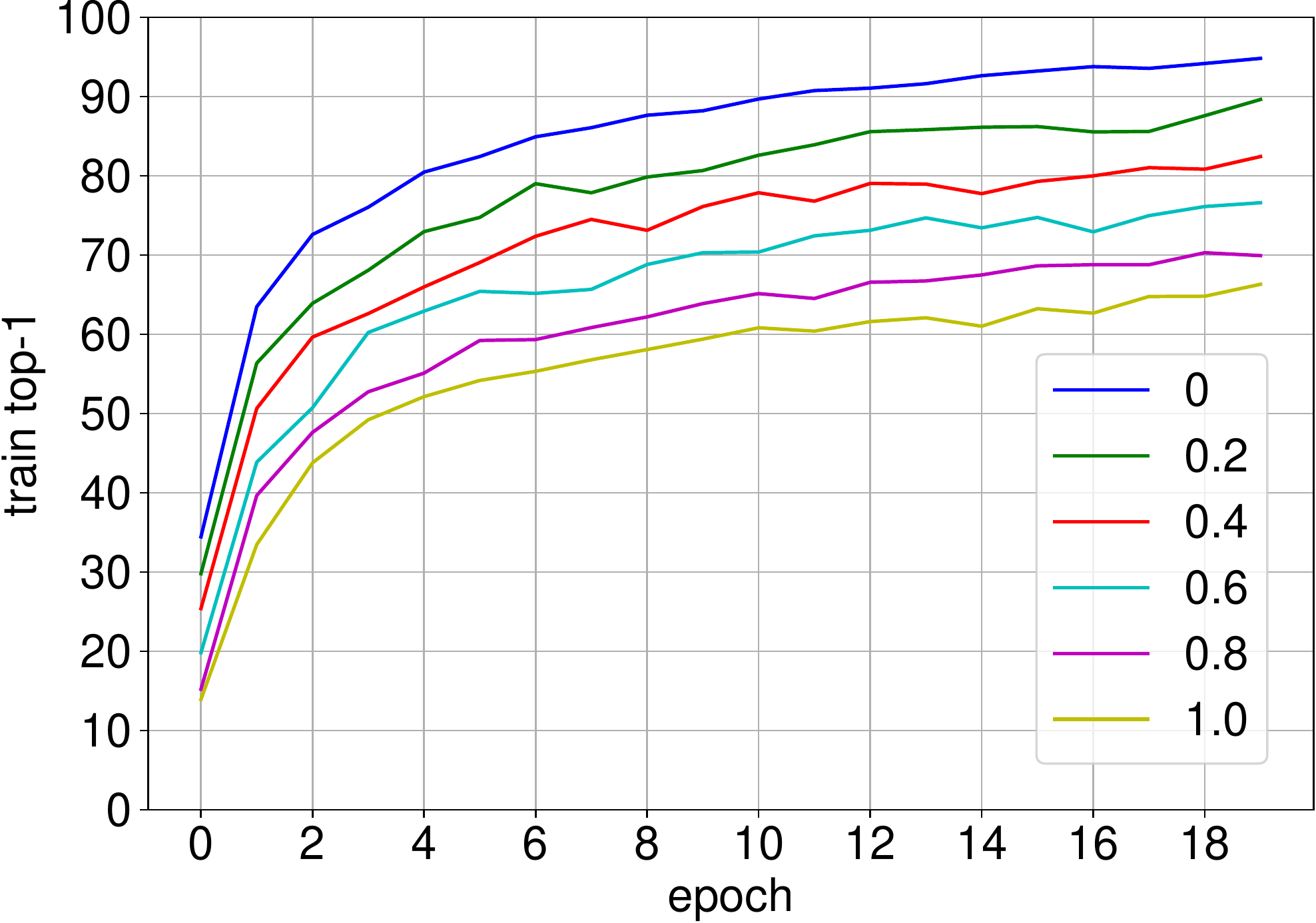}
    \subcaption{HMDB51}
    \label{fig:hmdb_ep_acc}
    \end{minipage}

    \caption{
    Performance of ObjectMix for different $p$.
    The top row shows the top-1 performance on the validation set and the bottom row shows that on the training set.
    }

    \label{fig:ObjectMix_graph}
\end{figure}

The performances over training epochs are shown in Figure \ref{fig:ObjectMix_graph}.
The validation results show that the performance is worse than the case with $p=0$ in the early stages of training, whereas it becomes equal or better as the training progresses, regardless of the value of $p$.
On the other hand, for $p=0$, the validation performance begins to deteriorate in the middle stage of the training even though the performance of the training set is consistently high and continues to increase,
indicating that overfitting occurs due to the lack of data augmentation.
On the other hand, the proposed method suppresses overfitting even with a small amount of augmentation with $p=0.2$.
The training performance degrades as $p$ increases, but this does not have much impact on the validation performance.

\subsection{Results of ObjectMix+or}

\begin{figure}[t]
    \centering
    
    \begin{minipage}[t]{0.49\linewidth}
    \centering
    \includegraphics[width=\linewidth]{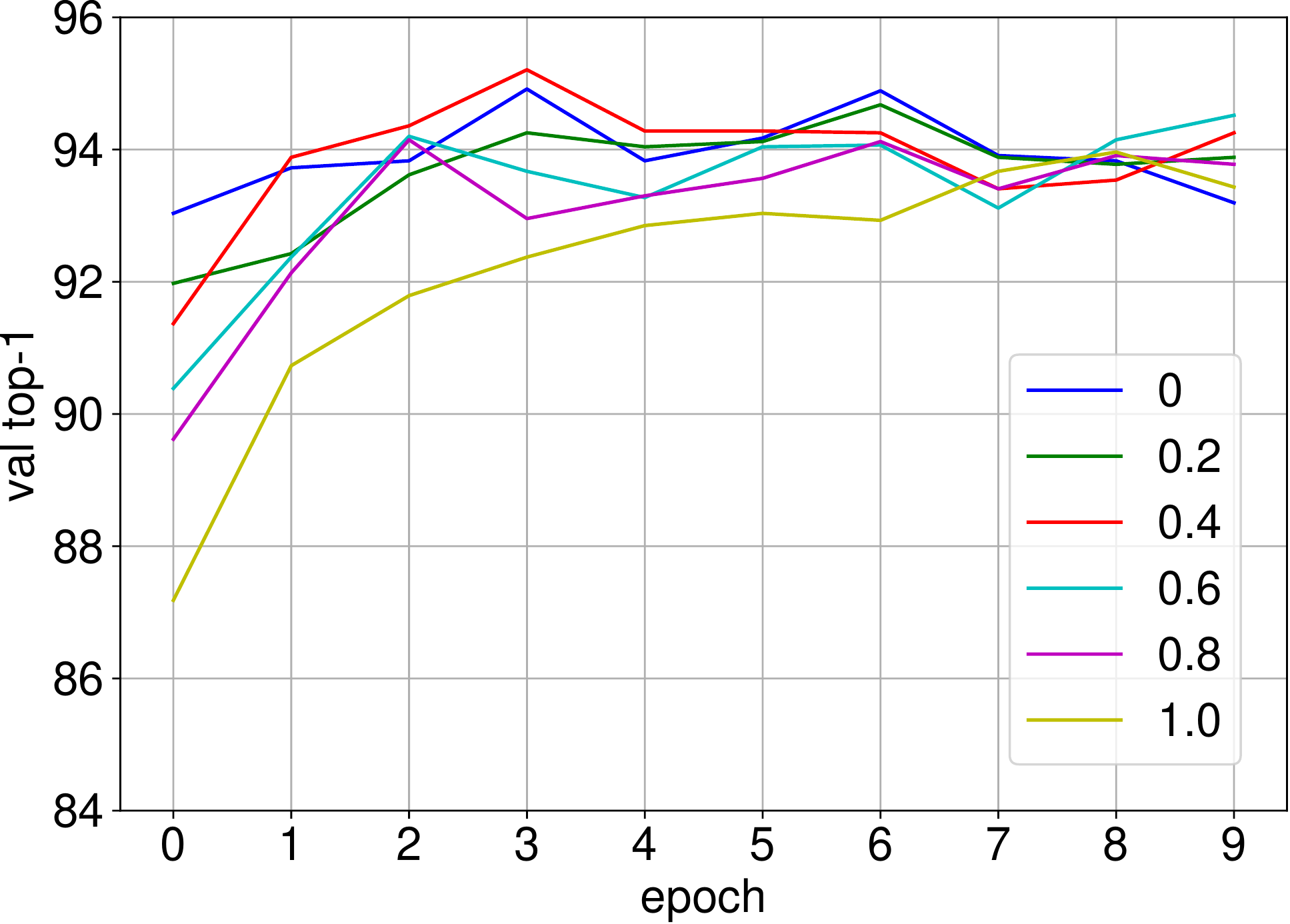}
    \end{minipage}
    \hfill
    \begin{minipage}[t]{0.49\linewidth}
    \centering
    \includegraphics[width=\linewidth]{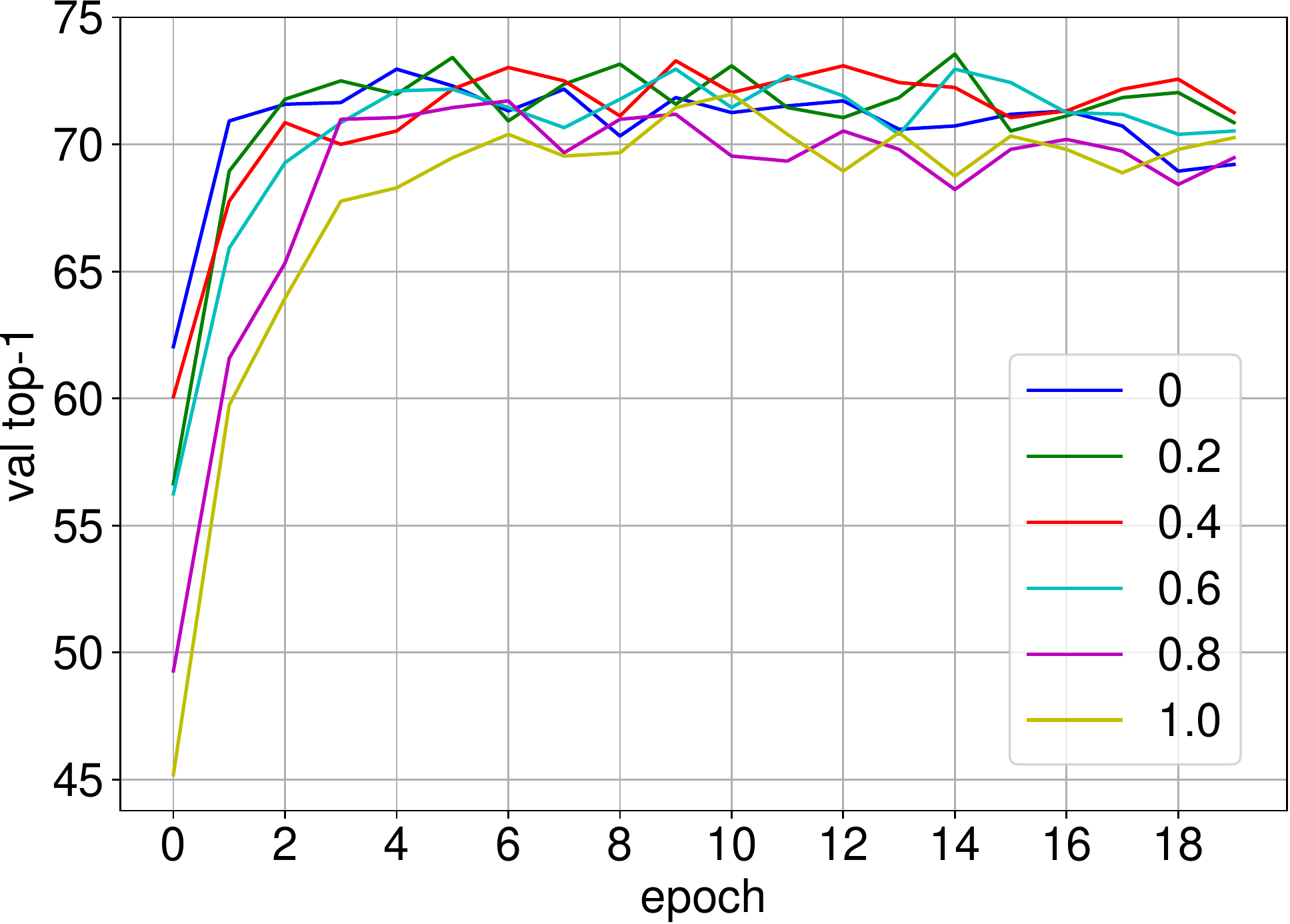}
    \end{minipage}

    \begin{minipage}[t]{0.49\linewidth}
    \centering
    \includegraphics[width=\linewidth]{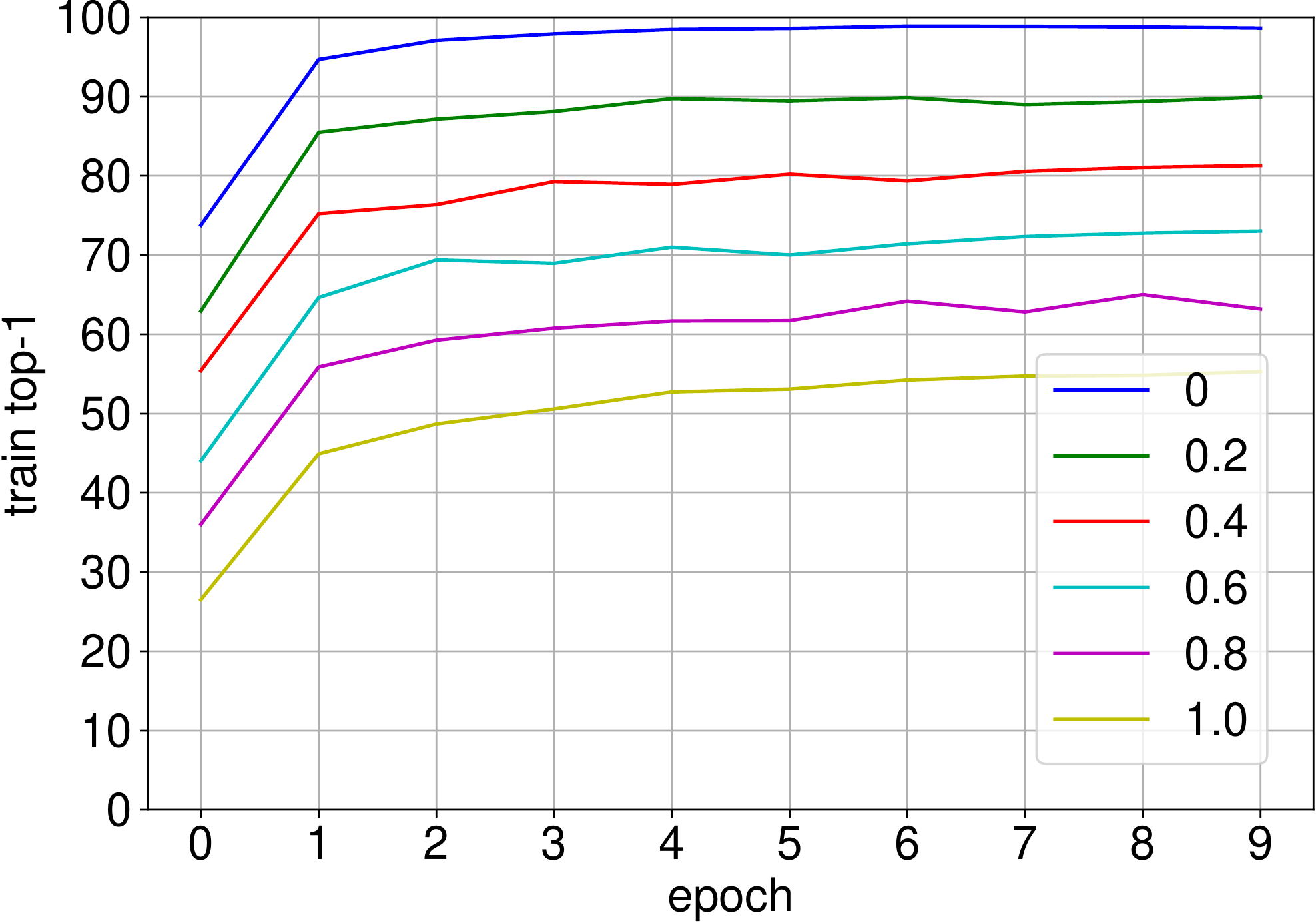}
    \subcaption{UCF101}
    \label{fig:ucf_or_acc_t}
    \end{minipage}
    \hfill
    \begin{minipage}[t]{0.49\linewidth}
    \centering
    \includegraphics[width=\linewidth]{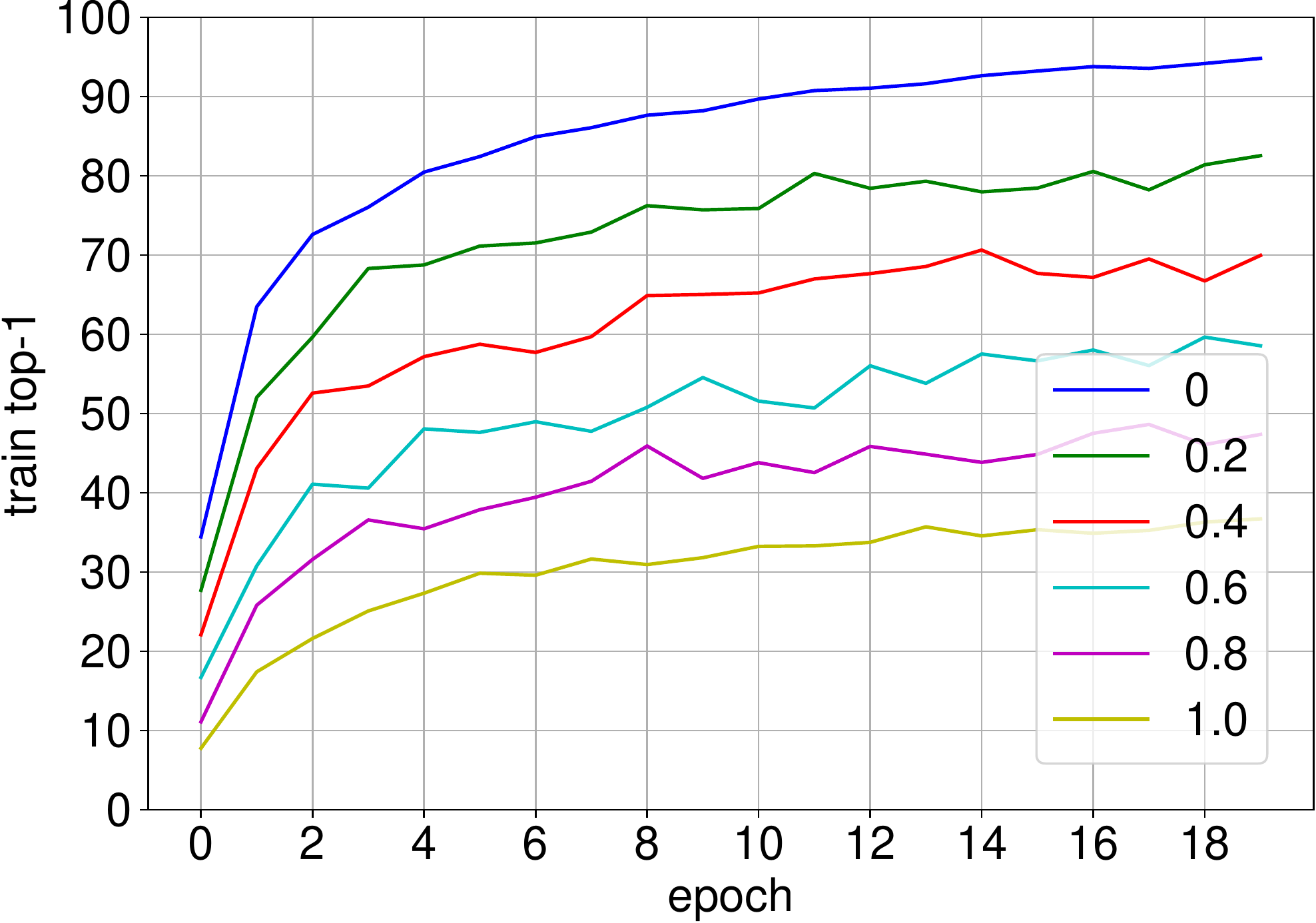}
    \subcaption{HMDB51}
    \label{fig:hmdb_or_acc_t}
    \end{minipage}

    \caption{
    Performance of ObjectMix+or for different $p$.
    }
    \label{fig:ObjectMix+or_graph}

\end{figure}

Next, we show results of ObjectMix+or in Table \ref{tab:ucf_or},
and the performances over training epochs in Figure \ref{fig:ObjectMix+or_graph}.
This result shows a similar trend of ObjectMix for both data sets;
the performance have a peak as $p$ increases.
Compared to ObjectMix, the performance on the training set is significantly lower,
indicating that ObjectMix+or plays a greater role in regularization as a data augmentation.
The validation performances are however similar (or slightly inferior) to those of ObjectMix,
which may shows that
frame-wise segmentation masks of ObjectMix might be enough even with failure makes in some frames.

\begin{table}[t]

\centering

\caption{The top-1 performance of ObjectMix+or (OM+or) on UCF101 and HMDB51 validation set.
}
\label{tab:ucf_or}
\label{tab:hmdb_or}

\scalebox{\tabscaleA}{
\begin{tabular}{c|cc|cc}
\multicolumn{1}{c|}{} & \multicolumn{2}{c|}{UCF101} & \multicolumn{2}{c}{HMDB51} \\
method (p)      & top-1 & top-5 & top-1 & top-5 \\ \hline
OM+or (0.0)  & $93.58 \pm  0.03$ & $99.23 \pm 0.01$ & $69.86 \pm  0.39$ & $91.89 \pm 0.26$ \\ \hline
OM+or (0.2)  & $94.08 \pm  0.18$ & $\textbf{99.54} \pm 0.07$ & $70.86 \pm  0.53$ & $92.49 \pm 0.28$ \\
OM+or (0.4)  & $94.19 \pm  0.09$ & $99.49 \pm 0.06$ & $\textbf{71.70} \pm  0.62$ & $92.56 \pm 0.22$ \\
OM+or (0.6)  & $\textbf{94.28} \pm  0.22$ & $99.31 \pm 0.06$ & $70.88 \pm  0.54$ & $\textbf{93.20} \pm 0.29$ \\
OM+or (0.8)  & $93.67 \pm  0.25$ & $99.22 \pm 0.05$ & $68.93 \pm  0.43$ & $91.75 \pm 0.25$ \\
OM+or (1.0)  & $93.25 \pm  0.23$ & $99.00 \pm 0.07$ & $69.88 \pm  0.45$ & $92.08 \pm 0.33$ \\
\end{tabular}
}

\end{table}

\subsection{Comparison with VideoMix}

\begin{figure}[t]
    \centering
    
    \begin{minipage}[t]{0.49\linewidth}
    \centering
    \includegraphics[width=\linewidth]{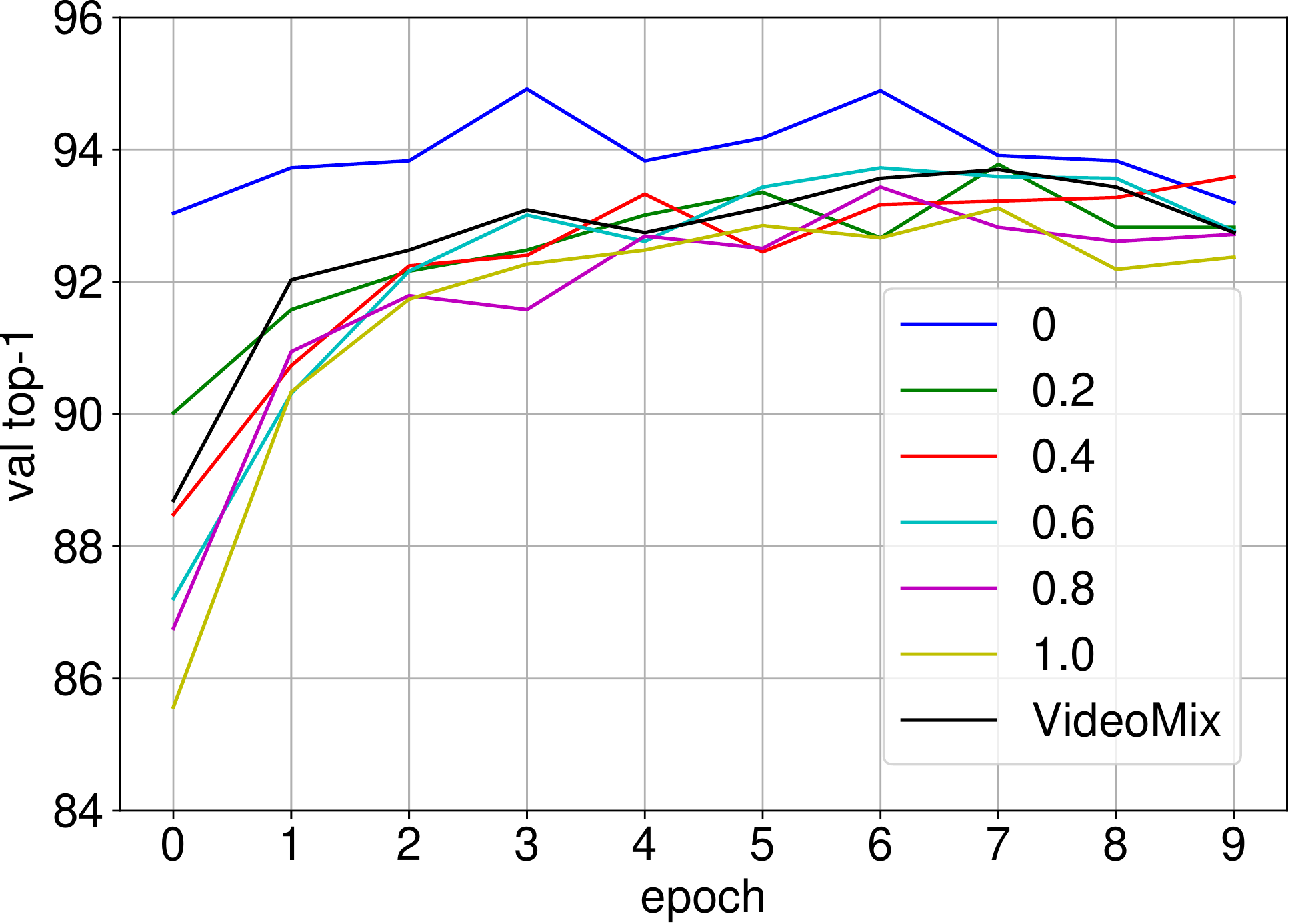}
    \end{minipage}
    \hfil
    \begin{minipage}[t]{0.49\linewidth}
    \centering
    \includegraphics[width=\linewidth]{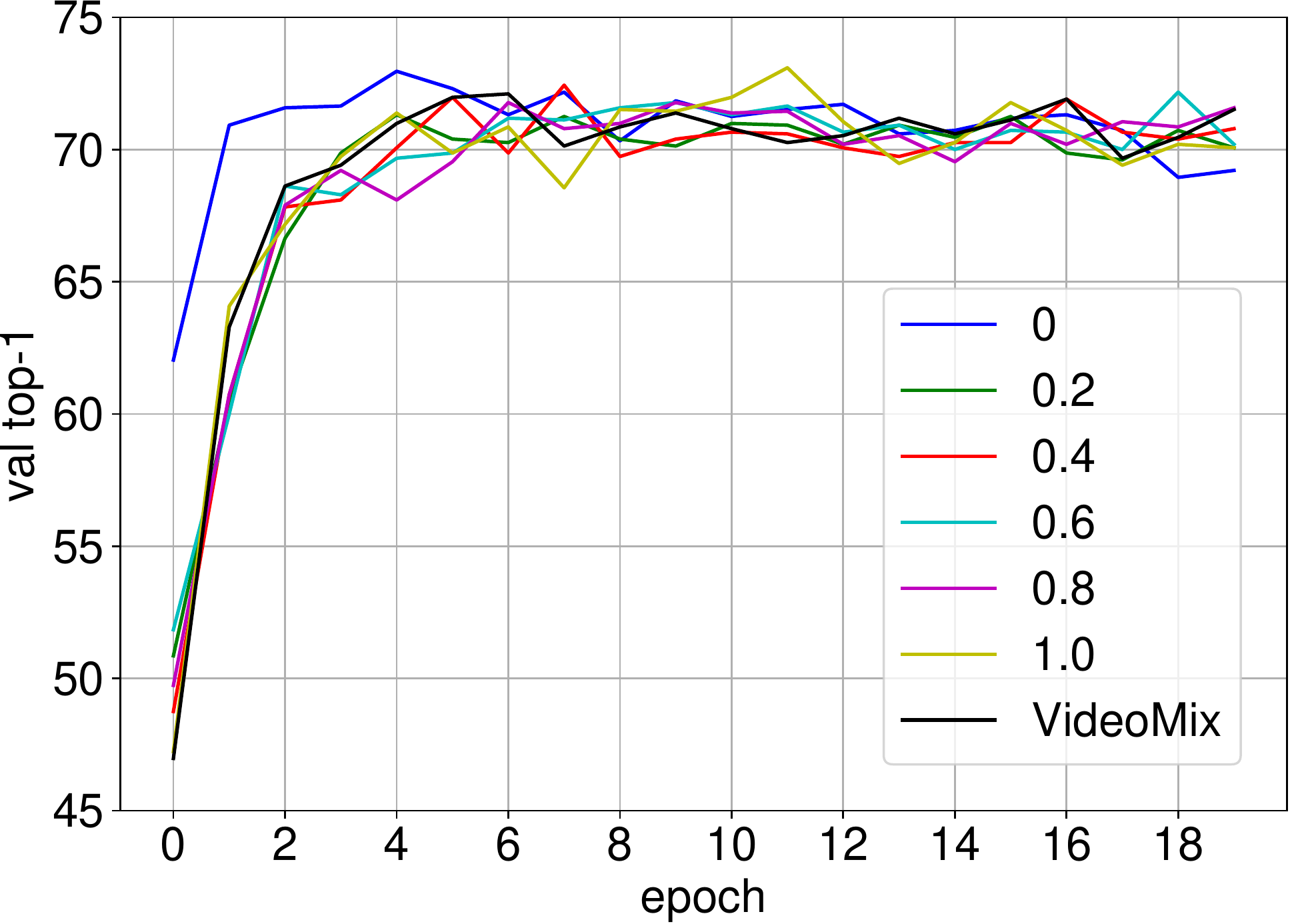}
    \end{minipage}

    \begin{minipage}[t]{0.49\linewidth}
    \centering
    \includegraphics[width=\linewidth]{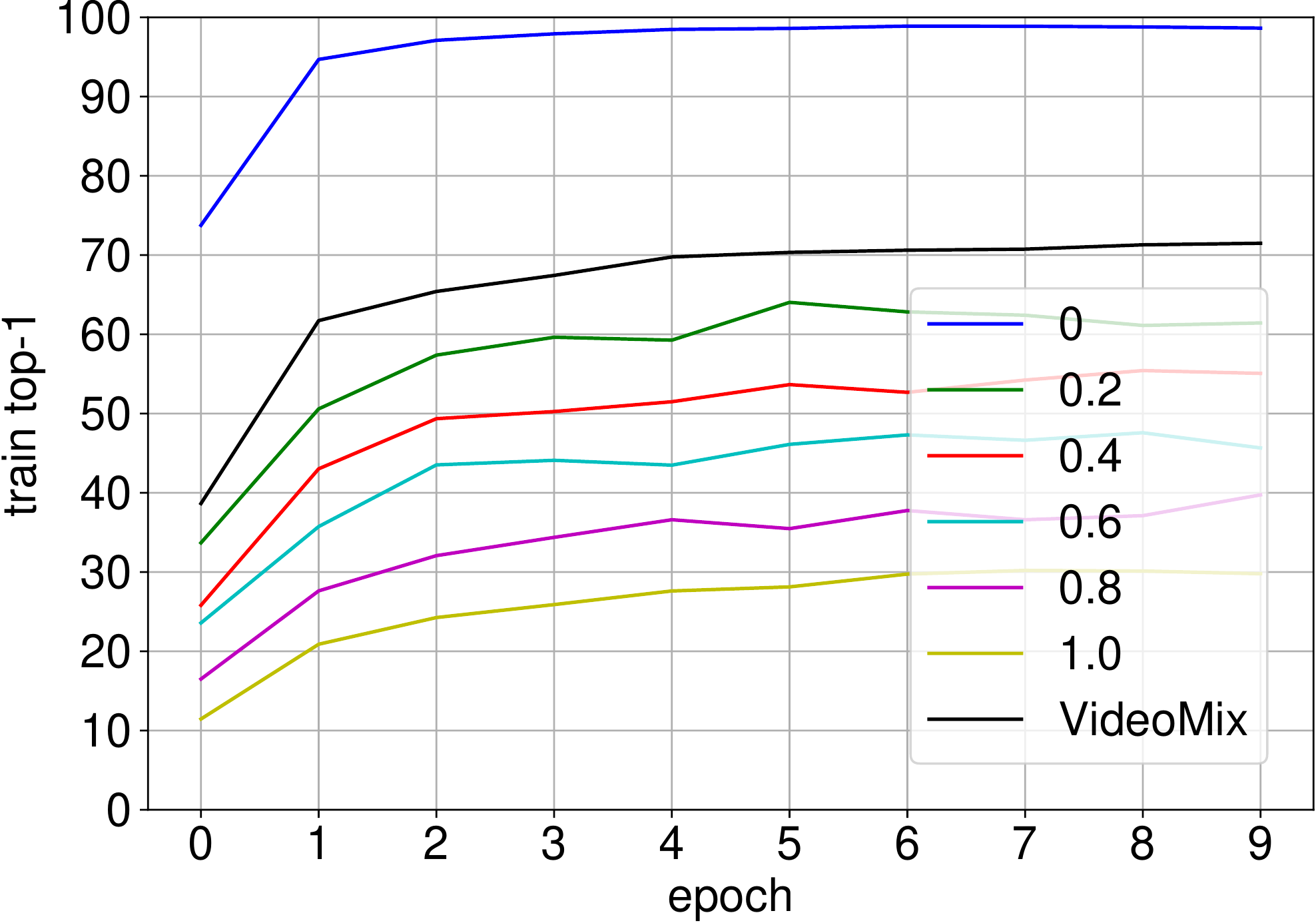}
    \subcaption{UCF101}
    \label{fig:ucf_mix_acc_t}
    \end{minipage}
    \hfill
    \begin{minipage}[t]{0.49\linewidth}
    \centering
    \includegraphics[width=\linewidth]{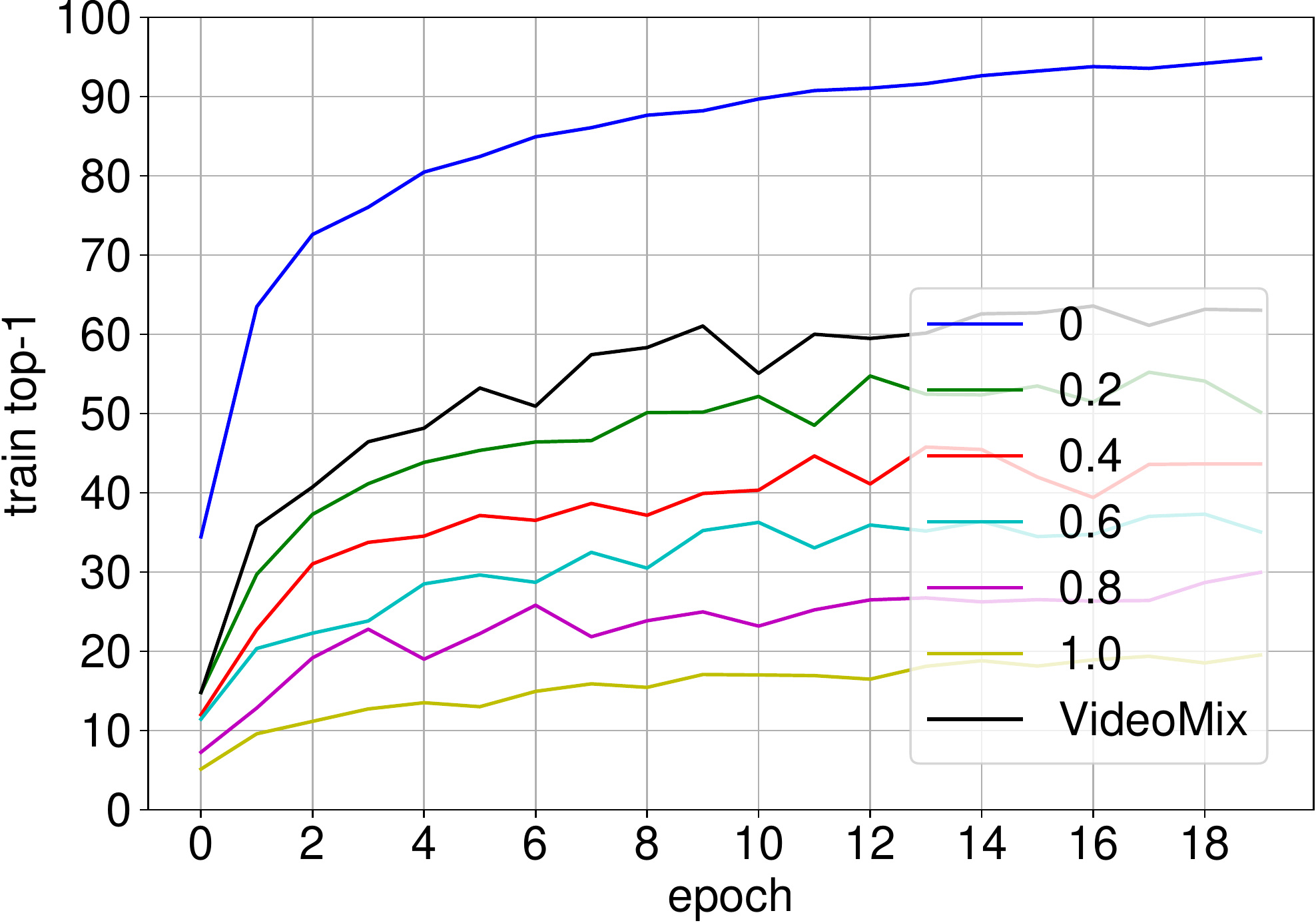}
    \subcaption{HMDB51}
    \label{fig:hmdb_mix_acc_t}
    \end{minipage}

    \caption{
    Performance comparisons of ObjectMix for different $p$ and VideoMix.
    }
    \label{fig:ObjectMix+or+mix_graph}
\end{figure}

Here we report the effect of using the proposed method in combination with VideoMix.
The settings followed the original paper \cite{VideoMix},
and a patch with center coordinates $(w_c, h_c)$ was sampled as follows;
\begin{align}
    \lambda &\sim \mathrm{Beta}(\alpha, \alpha), \quad \alpha = 1\\
    w_c &\sim \mathrm{Unif}(0, W), \quad W = 224 \\
    h_c &\sim \mathrm{Unif}(0, H), \quad H = 224 \\
    w_1 &= \max\left(0, w_c - \frac{W\sqrt{\lambda}}{2}\right), \quad
    w_2 = w_c + \frac{W\sqrt{\lambda}}{2} \\
    h_1 &= h_c - \frac{H\sqrt{\lambda}}{2}, \quad
    h_2 = h_c + \frac{H\sqrt{\lambda}}{2}.
\end{align}
We used S-VideoMix, which uses the same spatial patch across all frames,
and set the probability to $p=1$ according to the original paper.
The results are shown in Table \ref{tab:ucf_mix} and Figure \ref{fig:ObjectMix+or+mix_graph}.

The combination of ObjectMix and VideoMix shows a significant performance degradation compared to using ObjectMix alone.
One reason might be the fact that the size of the extracted mask region can bee too large by merging two regions from both ObjectMix and VideoMix.
A possible improvement is to search reasonable parameters of VideoMix, taking into account the mask size issue when used combined with ObjectMix. 

The performance of ObjectMix alone outperforms that of VideoMix alone shown here for any $p$ value,
indicating that the proposed method is more effective.

\begin{table}[t]

\centering
\caption{The top-1 performance of the combination of ObjectMix+or (OM+or) and VideoMix (VM) on UCF101 and HMDB51 validation set.
}
\label{tab:ucf_mix}
\label{tab:hmdb_mix}

\scalebox{\tabscaleB}{
\begin{tabular}{c|cc|cc}
\multicolumn{1}{c|}{} & \multicolumn{2}{c|}{UCF101} & \multicolumn{2}{c}{HMDB51} \\
method (p)      & top-1 & top-5 & top-1 & top-5 \\ \hline
OM+or (0.0)\phantom{+VM} & $93.58 \pm  0.03$ & $99.23 \pm 0.01$ & $69.86 \pm  0.39$ & $91.89 \pm 0.26$ \\ \hline
\hfill VM  & $93.34 \pm 0.07$ & $99.38 \pm 0.00$ &$70.90 \pm 0.34$ &$92.84 \pm 0.24$ \\ \hline
OM+or (0.2) +VM & $93.15 \pm  0.20$ & $99.26 \pm 0.06$ & $69.83 \pm  0.47$ & $92.10 \pm 0.35$ \\
OM+or (0.4) +VM & $\textbf{93.77} \pm  0.16$ & $\textbf{99.41} \pm 0.09$ & $70.49 \pm  0.54$ & $92.09 \pm 0.25$ \\
OM+or (0.6) +VM & $92.32 \pm  0.26$ & $99.07 \pm 0.06$ & $70.74 \pm  0.56$ & $\textbf{92.68} \pm 0.32$ \\
OM+or (0.8) +VM & $92.92 \pm  0.25$ & $99.19 \pm 0.09$ & $\textbf{71.35} \pm  0.29$ & $92.54 \pm 0.17$ \\
OM+or (1.0) +VM & $92.72 \pm  0.19$ & $99.04 \pm 0.08$ & $70.15 \pm  0.46$ & $92.52 \pm 0.24$ \\
\end{tabular}
}

\end{table}

\begin{table}[t]

\centering
\caption{Summary of the proposed method (fixed to $p=0.6$) and
comparisons of data augmentation (upper rows) with self-supervised methods (lower rows).}
\label{tab:sota_comparison}

\scalebox{\tabscaleA}{
\begin{tabular}{c|c|c}
 & UCF101 & HMDB51 \\
method  & top-1 & top-1 \\ \hline
no augmentation & $93.58 \pm 0.03$ & $69.86 \pm  0.39$ \\
OM (0.6)    & $95.12 \pm  0.15$ & $71.44 \pm  0.59$ \\
OM+or (0.6) & $94.28 \pm  0.22$ & $70.88 \pm  0.54$  \\
VideoMix \cite{VideoMix}        & 93.4\phantom{$0 \pm 0.00$} & 66.9\phantom{$0 \pm 0.00$} \\ \hline
CMD \cite{Huang_2021_CVPR}      & 85.7\phantom{$0 \pm 0.00$} & 54.0\phantom{$0 \pm 0.00$} \\
MoCo+BE \cite{Wang_2021_CVPR}   & 87.1\phantom{$0 \pm 0.00$} & 56.2\phantom{$0 \pm 0.00$} \\
CVRL \cite{Qian_2021_CVPR}      & 94.4\phantom{$0 \pm 0.00$} & 70.6\phantom{$0 \pm 0.00$} \\
VideoMAE \cite{VideoMAE} & 96.1\phantom{$0 \pm 0.00$} & 73.3\phantom{$0 \pm 0.00$} \\
$\rho$BYOL \cite{Feichtenhofer_2021_CVPR} & 96.3\phantom{$0 \pm 0.00$} & 75.0\phantom{$0 \pm 0.00$} \\
\end{tabular}
}

\end{table}

\section{Conclusion}

In this paper, we proposed ObjectMix, a data augmentation for action recognition, which uses object masks extracted from input video frames.
The proposed method differs from mix-based VideoMix; ObjectMix creates new videos by extracting objects rather than cutting and pasting random rectangles.
Experiments using UCF101 and HMDB51 have confirmed that the proposed method is effective to suppress overfitting.
The reasonable value of $p$ was about 0.5 for both datasets, which might be a good balance of the original and generated video samples,
and we will verify the results by using other datasets.
An obvious limitation of the proposed method is its computational cost.
While random rectangle cropping is nearly zero-cost,
applying instance segmentation is computationally expensive in both space and time,
hindering an efficient model training.
A less accurate but lighter model could be used to speed up mask extraction
because the accuracy of segmentation might have a small impact on the classification performance.
Therefore, a future work includes verification of the trade-off between performance and cost with such an efficient model.

Another topic of the future work is the comparison with self-supervised learning for video representation.
Table \ref{tab:sota_comparison} summarizes our results and compares the performance with the recent self-supervised methods.
This result shows that
self-supervised learning is becoming more effective than supervised learning with data augmentation.
Future work includes further investigation of how to combine the proposed segmentation-based augmentation method with self-supervised learning \cite{Kugelgen2021NeurIPS}.

\begin{acks}
This work was supported in part by
JSPS KAKENHI Grant Number JP22K12090.
\end{acks}

\bibliographystyle{ACM-Reference-Format}
\bibliography{mybib}


\begin{thebibliography}{41}


\ifx \showCODEN    \undefined \def \showCODEN     #1{\unskip}     \fi
\ifx \showDOI      \undefined \def \showDOI       #1{#1}\fi
\ifx \showISBNx    \undefined \def \showISBNx     #1{\unskip}     \fi
\ifx \showISBNxiii \undefined \def \showISBNxiii  #1{\unskip}     \fi
\ifx \showISSN     \undefined \def \showISSN      #1{\unskip}     \fi
\ifx \showLCCN     \undefined \def \showLCCN      #1{\unskip}     \fi
\ifx \shownote     \undefined \def \shownote      #1{#1}          \fi
\ifx \showarticletitle \undefined \def \showarticletitle #1{#1}   \fi
\ifx \showURL      \undefined \def \showURL       {\relax}        \fi
\providecommand\bibfield[2]{#2}
\providecommand\bibinfo[2]{#2}
\providecommand\natexlab[1]{#1}
\providecommand\showeprint[2][]{arXiv:#2}

\bibitem[\protect\citeauthoryear{Buslaev, Iglovikov, Khvedchenya, Parinov,
  Druzhinin, and Kalinin}{Buslaev et~al\mbox{.}}{2020}]%
        {Albumentations}
\bibfield{author}{\bibinfo{person}{Alexander Buslaev},
  \bibinfo{person}{Vladimir~I. Iglovikov}, \bibinfo{person}{Eugene
  Khvedchenya}, \bibinfo{person}{Alex Parinov}, \bibinfo{person}{Mikhail
  Druzhinin}, {and} \bibinfo{person}{Alexandr~A. Kalinin}.}
  \bibinfo{year}{2020}\natexlab{}.
\newblock \showarticletitle{Albumentations: Fast and Flexible Image
  Augmentations}.
\newblock \bibinfo{journal}{\emph{Information}} \bibinfo{volume}{11},
  \bibinfo{number}{2} (\bibinfo{year}{2020}).
\newblock
\showISSN{2078-2489}
\urldef\tempurl%
\url{https://doi.org/10.3390/info11020125}
\showDOI{\tempurl}


\bibitem[\protect\citeauthoryear{Butler, Wulff, Stanley, and Black}{Butler
  et~al\mbox{.}}{2012}]%
        {Sintel}
\bibfield{author}{\bibinfo{person}{D.~J. Butler}, \bibinfo{person}{J. Wulff},
  \bibinfo{person}{G.~B. Stanley}, {and} \bibinfo{person}{M.~J. Black}.}
  \bibinfo{year}{2012}\natexlab{}.
\newblock \showarticletitle{A naturalistic open source movie for optical flow
  evaluation}. In \bibinfo{booktitle}{\emph{European Conf. on Computer Vision
  (ECCV)}} \emph{(\bibinfo{series}{Part IV, LNCS 7577})},
  \bibfield{editor}{\bibinfo{person}{{A. Fitzgibbon et al. (Eds.)}}} (Ed.).
  \bibinfo{publisher}{Springer-Verlag}, \bibinfo{pages}{611--625}.
\newblock


\bibitem[\protect\citeauthoryear{Devries and Taylor}{Devries and
  Taylor}{2017}]%
        {DBLP:journals/corr/abs-1708-04552}
\bibfield{author}{\bibinfo{person}{Terrance Devries} {and}
  \bibinfo{person}{Graham~W. Taylor}.} \bibinfo{year}{2017}\natexlab{}.
\newblock \showarticletitle{Improved Regularization of Convolutional Neural
  Networks with Cutout}.
\newblock \bibinfo{journal}{\emph{CoRR}}  \bibinfo{volume}{abs/1708.04552}
  (\bibinfo{year}{2017}).
\newblock
\showeprint[arXiv]{1708.04552}
\urldef\tempurl%
\url{http://arxiv.org/abs/1708.04552}
\showURL{%
\tempurl}


\bibitem[\protect\citeauthoryear{Dosovitskiy, Fischer, Ilg, H{\"a}usser,
  Haz{\i}rba{\c{s}}, Golkov, v.d. Smagt, Cremers, and Brox}{Dosovitskiy
  et~al\mbox{.}}{2015}]%
        {FlyingChairs}
\bibfield{author}{\bibinfo{person}{A. Dosovitskiy}, \bibinfo{person}{P.
  Fischer}, \bibinfo{person}{E. Ilg}, \bibinfo{person}{P. H{\"a}usser},
  \bibinfo{person}{C. Haz{\i}rba{\c{s}}}, \bibinfo{person}{V. Golkov},
  \bibinfo{person}{P. v.d. Smagt}, \bibinfo{person}{D. Cremers}, {and}
  \bibinfo{person}{T. Brox}.} \bibinfo{year}{2015}\natexlab{}.
\newblock \showarticletitle{FlowNet: Learning Optical Flow with Convolutional
  Networks}. In \bibinfo{booktitle}{\emph{IEEE International Conference on
  Computer Vision (ICCV)}}.
\newblock
\urldef\tempurl%
\url{http://lmb.informatik.uni-freiburg.de/Publications/2015/DFIB15}
\showURL{%
\tempurl}


\bibitem[\protect\citeauthoryear{Dwibedi, Misra, and Hebert}{Dwibedi
  et~al\mbox{.}}{2017}]%
        {cutpaste}
\bibfield{author}{\bibinfo{person}{Debidatta Dwibedi}, \bibinfo{person}{Ishan
  Misra}, {and} \bibinfo{person}{Martial Hebert}.}
  \bibinfo{year}{2017}\natexlab{}.
\newblock \showarticletitle{Cut, Paste and Learn: Surprisingly Easy Synthesis
  for Instance Detection}. In \bibinfo{booktitle}{\emph{Proceedings of the IEEE
  International Conference on Computer Vision (ICCV)}}.
\newblock


\bibitem[\protect\citeauthoryear{Emami, Dong, Nejad-Davarani, and
  Glide-Hurst}{Emami et~al\mbox{.}}{2021}]%
        {SA-GAN_CT}
\bibfield{author}{\bibinfo{person}{Hajar Emami}, \bibinfo{person}{Ming Dong},
  \bibinfo{person}{Siamak~P. Nejad-Davarani}, {and} \bibinfo{person}{Carri~K.
  Glide-Hurst}.} \bibinfo{year}{2021}\natexlab{}.
\newblock \showarticletitle{SA-GAN: Structure-Aware GAN for Organ-Preserving
  Synthetic CT Generation}. In \bibinfo{booktitle}{\emph{Medical Image
  Computing and Computer Assisted Intervention -- MICCAI 2021}},
  \bibfield{editor}{\bibinfo{person}{Marleen de~Bruijne},
  \bibinfo{person}{Philippe~C. Cattin}, \bibinfo{person}{St{\'e}phane Cotin},
  \bibinfo{person}{Nicolas Padoy}, \bibinfo{person}{Stefanie Speidel},
  \bibinfo{person}{Yefeng Zheng}, {and} \bibinfo{person}{Caroline Essert}}
  (Eds.). \bibinfo{publisher}{Springer International Publishing},
  \bibinfo{address}{Cham}, \bibinfo{pages}{471--481}.
\newblock
\showISBNx{978-3-030-87231-1}


\bibitem[\protect\citeauthoryear{Feichtenhofer}{Feichtenhofer}{2020}]%
        {X3D}
\bibfield{author}{\bibinfo{person}{Christoph Feichtenhofer}.}
  \bibinfo{year}{2020}\natexlab{}.
\newblock \showarticletitle{X3D: Expanding Architectures for Efficient Video
  Recognition}. In \bibinfo{booktitle}{\emph{Proceedings of the IEEE/CVF
  Conference on Computer Vision and Pattern Recognition (CVPR)}}.
\newblock


\bibitem[\protect\citeauthoryear{Feichtenhofer, Fan, Malik, and
  He}{Feichtenhofer et~al\mbox{.}}{2019}]%
        {SlowFast}
\bibfield{author}{\bibinfo{person}{Christoph Feichtenhofer},
  \bibinfo{person}{Haoqi Fan}, \bibinfo{person}{Jitendra Malik}, {and}
  \bibinfo{person}{Kaiming He}.} \bibinfo{year}{2019}\natexlab{}.
\newblock \showarticletitle{SlowFast Networks for Video Recognition}. In
  \bibinfo{booktitle}{\emph{Proceedings of the IEEE/CVF International
  Conference on Computer Vision (ICCV)}}.
\newblock


\bibitem[\protect\citeauthoryear{Feichtenhofer, Fan, Xiong, Girshick, and
  He}{Feichtenhofer et~al\mbox{.}}{2021}]%
        {Feichtenhofer_2021_CVPR}
\bibfield{author}{\bibinfo{person}{Christoph Feichtenhofer},
  \bibinfo{person}{Haoqi Fan}, \bibinfo{person}{Bo Xiong},
  \bibinfo{person}{Ross Girshick}, {and} \bibinfo{person}{Kaiming He}.}
  \bibinfo{year}{2021}\natexlab{}.
\newblock \showarticletitle{A Large-Scale Study on Unsupervised Spatiotemporal
  Representation Learning}. In \bibinfo{booktitle}{\emph{Proceedings of the
  IEEE/CVF Conference on Computer Vision and Pattern Recognition (CVPR)}}.
  \bibinfo{pages}{3299--3309}.
\newblock


\bibitem[\protect\citeauthoryear{Ghiasi, Cui, Srinivas, Qian, Lin, Cubuk, Le,
  and Zoph}{Ghiasi et~al\mbox{.}}{2021}]%
        {CopyPaste}
\bibfield{author}{\bibinfo{person}{Golnaz Ghiasi}, \bibinfo{person}{Yin Cui},
  \bibinfo{person}{Aravind Srinivas}, \bibinfo{person}{Rui Qian},
  \bibinfo{person}{Tsung-Yi Lin}, \bibinfo{person}{Ekin~D. Cubuk},
  \bibinfo{person}{Quoc~V. Le}, {and} \bibinfo{person}{Barret Zoph}.}
  \bibinfo{year}{2021}\natexlab{}.
\newblock \showarticletitle{Simple Copy-Paste Is a Strong Data Augmentation
  Method for Instance Segmentation}. In \bibinfo{booktitle}{\emph{Proceedings
  of the IEEE/CVF Conference on Computer Vision and Pattern Recognition
  (CVPR)}}. \bibinfo{pages}{2918--2928}.
\newblock


\bibitem[\protect\citeauthoryear{Goodfellow, Pouget-Abadie, Mirza, Xu,
  Warde-Farley, Ozair, Courville, and Bengio}{Goodfellow et~al\mbox{.}}{2014}]%
        {GAN}
\bibfield{author}{\bibinfo{person}{Ian Goodfellow}, \bibinfo{person}{Jean
  Pouget-Abadie}, \bibinfo{person}{Mehdi Mirza}, \bibinfo{person}{Bing Xu},
  \bibinfo{person}{David Warde-Farley}, \bibinfo{person}{Sherjil Ozair},
  \bibinfo{person}{Aaron Courville}, {and} \bibinfo{person}{Yoshua Bengio}.}
  \bibinfo{year}{2014}\natexlab{}.
\newblock \showarticletitle{Generative Adversarial Nets}. In
  \bibinfo{booktitle}{\emph{Advances in Neural Information Processing
  Systems}}, \bibfield{editor}{\bibinfo{person}{Z.~Ghahramani},
  \bibinfo{person}{M.~Welling}, \bibinfo{person}{C.~Cortes},
  \bibinfo{person}{N.~Lawrence}, {and} \bibinfo{person}{K.~Q. Weinberger}}
  (Eds.), Vol.~\bibinfo{volume}{27}. \bibinfo{publisher}{Curran Associates,
  Inc.}
\newblock
\urldef\tempurl%
\url{https://proceedings.neurips.cc/paper/2014/file/5ca3e9b122f61f8f06494c97b1afccf3-Paper.pdf}
\showURL{%
\tempurl}


\bibitem[\protect\citeauthoryear{Goyal, Ebrahimi~Kahou, Michalski, Materzynska,
  Westphal, Kim, Haenel, Fruend, Yianilos, Mueller-Freitag, Hoppe, Thurau, Bax,
  and Memisevic}{Goyal et~al\mbox{.}}{2017}]%
        {SSv2}
\bibfield{author}{\bibinfo{person}{Raghav Goyal}, \bibinfo{person}{Samira
  Ebrahimi~Kahou}, \bibinfo{person}{Vincent Michalski}, \bibinfo{person}{Joanna
  Materzynska}, \bibinfo{person}{Susanne Westphal}, \bibinfo{person}{Heuna
  Kim}, \bibinfo{person}{Valentin Haenel}, \bibinfo{person}{Ingo Fruend},
  \bibinfo{person}{Peter Yianilos}, \bibinfo{person}{Moritz Mueller-Freitag},
  \bibinfo{person}{Florian Hoppe}, \bibinfo{person}{Christian Thurau},
  \bibinfo{person}{Ingo Bax}, {and} \bibinfo{person}{Roland Memisevic}.}
  \bibinfo{year}{2017}\natexlab{}.
\newblock \showarticletitle{The "Something Something" Video Database for
  Learning and Evaluating Visual Common Sense}. In
  \bibinfo{booktitle}{\emph{Proceedings of the IEEE International Conference on
  Computer Vision (ICCV)}}.
\newblock


\bibitem[\protect\citeauthoryear{Hara, Kataoka, and Satoh}{Hara
  et~al\mbox{.}}{2018}]%
        {3Dresnet}
\bibfield{author}{\bibinfo{person}{Kensho Hara}, \bibinfo{person}{Hirokatsu
  Kataoka}, {and} \bibinfo{person}{Yutaka Satoh}.}
  \bibinfo{year}{2018}\natexlab{}.
\newblock \showarticletitle{Can Spatiotemporal 3D CNNs Retrace the History of
  2D CNNs and ImageNet?}. In \bibinfo{booktitle}{\emph{Proceedings of the IEEE
  Conference on Computer Vision and Pattern Recognition (CVPR)}}.
  \bibinfo{pages}{6546--6555}.
\newblock


\bibitem[\protect\citeauthoryear{Huang, Liu, Wang, Pan, Xu, and Jin}{Huang
  et~al\mbox{.}}{2021}]%
        {Huang_2021_CVPR}
\bibfield{author}{\bibinfo{person}{Lianghua Huang}, \bibinfo{person}{Yu Liu},
  \bibinfo{person}{Bin Wang}, \bibinfo{person}{Pan Pan},
  \bibinfo{person}{Yinghui Xu}, {and} \bibinfo{person}{Rong Jin}.}
  \bibinfo{year}{2021}\natexlab{}.
\newblock \showarticletitle{Self-Supervised Video Representation Learning by
  Context and Motion Decoupling}. In \bibinfo{booktitle}{\emph{Proceedings of
  the IEEE/CVF Conference on Computer Vision and Pattern Recognition (CVPR)}}.
  \bibinfo{pages}{13886--13895}.
\newblock


\bibitem[\protect\citeauthoryear{Hutchinson and Gadepally}{Hutchinson and
  Gadepally}{2021}]%
        {Video_Action_Understanding2021}
\bibfield{author}{\bibinfo{person}{Matthew~S. Hutchinson} {and}
  \bibinfo{person}{Vijay~N. Gadepally}.} \bibinfo{year}{2021}\natexlab{}.
\newblock \showarticletitle{Video Action Understanding}.
\newblock \bibinfo{journal}{\emph{IEEE Access}}  \bibinfo{volume}{9}
  (\bibinfo{year}{2021}), \bibinfo{pages}{134611--134637}.
\newblock
\urldef\tempurl%
\url{https://doi.org/10.1109/ACCESS.2021.3115476}
\showDOI{\tempurl}


\bibitem[\protect\citeauthoryear{Ishii and Yamashita}{Ishii and
  Yamashita}{2021}]%
        {CutDepth}
\bibfield{author}{\bibinfo{person}{Yasunori Ishii} {and}
  \bibinfo{person}{Takayoshi Yamashita}.} \bibinfo{year}{2021}\natexlab{}.
\newblock \showarticletitle{CutDepth: Edge-aware Data Augmentation in Depth
  Estimation}.
\newblock \bibinfo{journal}{\emph{CoRR}}  \bibinfo{volume}{abs/2107.07684}
  (\bibinfo{year}{2021}).
\newblock
\showeprint[arXiv]{2107.07684}
\urldef\tempurl%
\url{https://arxiv.org/abs/2107.07684}
\showURL{%
\tempurl}


\bibitem[\protect\citeauthoryear{Jung, Wada, Crall, Tanaka, Graving, Reinders,
  Yadav, Banerjee, Vecsei, Kraft, Rui, Borovec, Vallentin, Zhydenko, Pfeiffer,
  Cook, Fernández, De~Rainville, Weng, Ayala-Acevedo, Meudec, Laporte,
  et~al\mbox{.}}{Jung et~al\mbox{.}}{2020}]%
        {imgaug}
\bibfield{author}{\bibinfo{person}{Alexander~B. Jung}, \bibinfo{person}{Kentaro
  Wada}, \bibinfo{person}{Jon Crall}, \bibinfo{person}{Satoshi Tanaka},
  \bibinfo{person}{Jake Graving}, \bibinfo{person}{Christoph Reinders},
  \bibinfo{person}{Sarthak Yadav}, \bibinfo{person}{Joy Banerjee},
  \bibinfo{person}{Gábor Vecsei}, \bibinfo{person}{Adam Kraft},
  \bibinfo{person}{Zheng Rui}, \bibinfo{person}{Jirka Borovec},
  \bibinfo{person}{Christian Vallentin}, \bibinfo{person}{Semen Zhydenko},
  \bibinfo{person}{Kilian Pfeiffer}, \bibinfo{person}{Ben Cook},
  \bibinfo{person}{Ismael Fernández}, \bibinfo{person}{François-Michel
  De~Rainville}, \bibinfo{person}{Chi-Hung Weng}, \bibinfo{person}{Abner
  Ayala-Acevedo}, \bibinfo{person}{Raphael Meudec}, \bibinfo{person}{Matias
  Laporte}, {et~al\mbox{.}}} \bibinfo{year}{2020}\natexlab{}.
\newblock \bibinfo{title}{{imgaug}}.
\newblock \bibinfo{howpublished}{\url{https://github.com/aleju/imgaug}}.
\newblock
\newblock
\shownote{Online; accessed 01-Feb-2020}.


\bibitem[\protect\citeauthoryear{Jung, Luna, and Park}{Jung
  et~al\mbox{.}}{2021}]%
        {cGAN_Medical}
\bibfield{author}{\bibinfo{person}{Euijin Jung}, \bibinfo{person}{Miguel Luna},
  {and} \bibinfo{person}{Sang~Hyun Park}.} \bibinfo{year}{2021}\natexlab{}.
\newblock \showarticletitle{Conditional GAN with an Attention-Based Generator
  and a 3D Discriminator for 3D Medical Image Generation}. In
  \bibinfo{booktitle}{\emph{Medical Image Computing and Computer Assisted
  Intervention -- MICCAI 2021}}, \bibfield{editor}{\bibinfo{person}{Marleen
  de~Bruijne}, \bibinfo{person}{Philippe~C. Cattin},
  \bibinfo{person}{St{\'e}phane Cotin}, \bibinfo{person}{Nicolas Padoy},
  \bibinfo{person}{Stefanie Speidel}, \bibinfo{person}{Yefeng Zheng}, {and}
  \bibinfo{person}{Caroline Essert}} (Eds.). \bibinfo{publisher}{Springer
  International Publishing}, \bibinfo{address}{Cham},
  \bibinfo{pages}{318--328}.
\newblock
\showISBNx{978-3-030-87231-1}


\bibitem[\protect\citeauthoryear{Kay, Carreira, Simonyan, Zhang, Hillier,
  Vijayanarasimhan, Viola, Green, Back, Natsev, Suleyman, and Zisserman}{Kay
  et~al\mbox{.}}{2017}]%
        {DBLP:journals/corr/Kinetics}
\bibfield{author}{\bibinfo{person}{Will Kay}, \bibinfo{person}{Jo{\~{a}}o
  Carreira}, \bibinfo{person}{Karen Simonyan}, \bibinfo{person}{Brian Zhang},
  \bibinfo{person}{Chloe Hillier}, \bibinfo{person}{Sudheendra
  Vijayanarasimhan}, \bibinfo{person}{Fabio Viola}, \bibinfo{person}{Tim
  Green}, \bibinfo{person}{Trevor Back}, \bibinfo{person}{Paul Natsev},
  \bibinfo{person}{Mustafa Suleyman}, {and} \bibinfo{person}{Andrew
  Zisserman}.} \bibinfo{year}{2017}\natexlab{}.
\newblock \showarticletitle{The Kinetics Human Action Video Dataset}.
\newblock \bibinfo{journal}{\emph{CoRR}}  \bibinfo{volume}{abs/1705.06950}
  (\bibinfo{year}{2017}).
\newblock
\showeprint[arXiv]{1705.06950}
\urldef\tempurl%
\url{http://arxiv.org/abs/1705.06950}
\showURL{%
\tempurl}


\bibitem[\protect\citeauthoryear{Khalifa, Loey, and Mirjalili}{Khalifa
  et~al\mbox{.}}{2022}]%
        {DBLP:journals/air/KhalifaLM22}
\bibfield{author}{\bibinfo{person}{Nour Eldeen~Mahmoud Khalifa},
  \bibinfo{person}{Mohamed Loey}, {and} \bibinfo{person}{Seyedali Mirjalili}.}
  \bibinfo{year}{2022}\natexlab{}.
\newblock \showarticletitle{A comprehensive survey of recent trends in deep
  learning for digital images augmentation}.
\newblock \bibinfo{journal}{\emph{Artificial Intelligence Review}}
  \bibinfo{volume}{55}, \bibinfo{number}{3} (\bibinfo{year}{2022}),
  \bibinfo{pages}{2351--2377}.
\newblock
\urldef\tempurl%
\url{https://doi.org/10.1007/s10462-021-10066-4}
\showDOI{\tempurl}


\bibitem[\protect\citeauthoryear{Lin, Maire, Belongie, Bourdev, Girshick, Hays,
  Perona, Ramanan, Doll{\'{a}}r, and Zitnick}{Lin et~al\mbox{.}}{2014}]%
        {MSCOCO}
\bibfield{author}{\bibinfo{person}{Tsung{-}Yi Lin}, \bibinfo{person}{Michael
  Maire}, \bibinfo{person}{Serge~J. Belongie}, \bibinfo{person}{Lubomir~D.
  Bourdev}, \bibinfo{person}{Ross~B. Girshick}, \bibinfo{person}{James Hays},
  \bibinfo{person}{Pietro Perona}, \bibinfo{person}{Deva Ramanan},
  \bibinfo{person}{Piotr Doll{\'{a}}r}, {and} \bibinfo{person}{C.~Lawrence
  Zitnick}.} \bibinfo{year}{2014}\natexlab{}.
\newblock \showarticletitle{Microsoft {COCO:} Common Objects in Context}.
\newblock \bibinfo{journal}{\emph{CoRR}}  \bibinfo{volume}{abs/1405.0312}
  (\bibinfo{year}{2014}).
\newblock
\showeprint[arXiv]{1405.0312}
\urldef\tempurl%
\url{http://arxiv.org/abs/1405.0312}
\showURL{%
\tempurl}


\bibitem[\protect\citeauthoryear{Mayer, Ilg, H{\"a}usser, Fischer, Cremers,
  Dosovitskiy, and Brox}{Mayer et~al\mbox{.}}{2016}]%
        {Flyingthings3D}
\bibfield{author}{\bibinfo{person}{N. Mayer}, \bibinfo{person}{E. Ilg},
  \bibinfo{person}{P. H{\"a}usser}, \bibinfo{person}{P. Fischer},
  \bibinfo{person}{D. Cremers}, \bibinfo{person}{A. Dosovitskiy}, {and}
  \bibinfo{person}{T. Brox}.} \bibinfo{year}{2016}\natexlab{}.
\newblock \showarticletitle{A Large Dataset to Train Convolutional Networks for
  Disparity, Optical Flow, and Scene Flow Estimation}. In
  \bibinfo{booktitle}{\emph{IEEE International Conference on Computer Vision
  and Pattern Recognition (CVPR)}}.
\newblock
\urldef\tempurl%
\url{http://lmb.informatik.uni-freiburg.de/Publications/2016/MIFDB16}
\showURL{%
\tempurl}
\newblock
\shownote{arXiv:1512.02134}.


\bibitem[\protect\citeauthoryear{Pang, Lin, Qin, and Chen}{Pang
  et~al\mbox{.}}{2021}]%
        {9528943}
\bibfield{author}{\bibinfo{person}{Yingxue Pang}, \bibinfo{person}{Jianxin
  Lin}, \bibinfo{person}{Tao Qin}, {and} \bibinfo{person}{Zhibo Chen}.}
  \bibinfo{year}{2021}\natexlab{}.
\newblock \showarticletitle{Image-to-Image Translation: Methods and
  Applications}.
\newblock \bibinfo{journal}{\emph{IEEE Transactions on Multimedia}}
  (\bibinfo{year}{2021}), \bibinfo{pages}{1--1}.
\newblock
\urldef\tempurl%
\url{https://doi.org/10.1109/TMM.2021.3109419}
\showDOI{\tempurl}


\bibitem[\protect\citeauthoryear{Qian, Meng, Gong, Yang, Wang, Belongie, and
  Cui}{Qian et~al\mbox{.}}{2021}]%
        {Qian_2021_CVPR}
\bibfield{author}{\bibinfo{person}{Rui Qian}, \bibinfo{person}{Tianjian Meng},
  \bibinfo{person}{Boqing Gong}, \bibinfo{person}{Ming-Hsuan Yang},
  \bibinfo{person}{Huisheng Wang}, \bibinfo{person}{Serge Belongie}, {and}
  \bibinfo{person}{Yin Cui}.} \bibinfo{year}{2021}\natexlab{}.
\newblock \showarticletitle{Spatiotemporal Contrastive Video Representation
  Learning}. In \bibinfo{booktitle}{\emph{Proceedings of the IEEE/CVF
  Conference on Computer Vision and Pattern Recognition (CVPR)}}.
  \bibinfo{pages}{6964--6974}.
\newblock


\bibitem[\protect\citeauthoryear{Shorten and Khoshgoftaar}{Shorten and
  Khoshgoftaar}{2019}]%
        {DBLP:journals/jbd/ShortenK19}
\bibfield{author}{\bibinfo{person}{Connor Shorten} {and}
  \bibinfo{person}{Taghi~M. Khoshgoftaar}.} \bibinfo{year}{2019}\natexlab{}.
\newblock \showarticletitle{A survey on Image Data Augmentation for Deep
  Learning}.
\newblock \bibinfo{journal}{\emph{Journal of Big Data}}  \bibinfo{volume}{6}
  (\bibinfo{year}{2019}), \bibinfo{pages}{60}.
\newblock
\urldef\tempurl%
\url{https://doi.org/10.1186/s40537-019-0197-0}
\showDOI{\tempurl}


\bibitem[\protect\citeauthoryear{Simonyan and Zisserman}{Simonyan and
  Zisserman}{2014}]%
        {TwoStream}
\bibfield{author}{\bibinfo{person}{Karen Simonyan} {and}
  \bibinfo{person}{Andrew Zisserman}.} \bibinfo{year}{2014}\natexlab{}.
\newblock \showarticletitle{Two-Stream Convolutional Networks for Action
  Recognition in Videos}. In \bibinfo{booktitle}{\emph{Advances in Neural
  Information Processing Systems}},
  \bibfield{editor}{\bibinfo{person}{Z.~Ghahramani},
  \bibinfo{person}{M.~Welling}, \bibinfo{person}{C.~Cortes},
  \bibinfo{person}{N.~Lawrence}, {and} \bibinfo{person}{K.~Q. Weinberger}}
  (Eds.), Vol.~\bibinfo{volume}{27}. \bibinfo{publisher}{Curran Associates,
  Inc.}
\newblock
\urldef\tempurl%
\url{https://proceedings.neurips.cc/paper/2014/file/00ec53c4682d36f5c4359f4ae7bd7ba1-Paper.pdf}
\showURL{%
\tempurl}


\bibitem[\protect\citeauthoryear{Soomro, Zamir, and Shah}{Soomro
  et~al\mbox{.}}{2012}]%
        {DBLP:ucf101}
\bibfield{author}{\bibinfo{person}{Khurram Soomro},
  \bibinfo{person}{Amir~Roshan Zamir}, {and} \bibinfo{person}{Mubarak Shah}.}
  \bibinfo{year}{2012}\natexlab{}.
\newblock \showarticletitle{{UCF101:} {A} Dataset of 101 Human Actions Classes
  From Videos in The Wild}.
\newblock \bibinfo{journal}{\emph{CoRR}}  \bibinfo{volume}{abs/1212.0402}
  (\bibinfo{year}{2012}).
\newblock
\showeprint[arXiv]{1212.0402}
\urldef\tempurl%
\url{http://arxiv.org/abs/1212.0402}
\showURL{%
\tempurl}


\bibitem[\protect\citeauthoryear{Tian, Ren, Chai, Olszewski, Peng, Metaxas, and
  Tulyakov}{Tian et~al\mbox{.}}{2021}]%
        {GAN_video}
\bibfield{author}{\bibinfo{person}{Yu Tian}, \bibinfo{person}{Jian Ren},
  \bibinfo{person}{Menglei Chai}, \bibinfo{person}{Kyle Olszewski},
  \bibinfo{person}{Xi Peng}, \bibinfo{person}{Dimitris~N. Metaxas}, {and}
  \bibinfo{person}{Sergey Tulyakov}.} \bibinfo{year}{2021}\natexlab{}.
\newblock \showarticletitle{A Good Image Generator Is What You Need for
  High-Resolution Video Synthesis}. In \bibinfo{booktitle}{\emph{International
  Conference on Learning Representations}}.
\newblock
\urldef\tempurl%
\url{https://openreview.net/forum?id=6puCSjH3hwA}
\showURL{%
\tempurl}


\bibitem[\protect\citeauthoryear{Tong, Song, Wang, and Wang}{Tong
  et~al\mbox{.}}{2022}]%
        {VideoMAE}
\bibfield{author}{\bibinfo{person}{Zhan Tong}, \bibinfo{person}{Yibing Song},
  \bibinfo{person}{Jue Wang}, {and} \bibinfo{person}{Limin Wang}.}
  \bibinfo{year}{2022}\natexlab{}.
\newblock \showarticletitle{VideoMAE: Masked Autoencoders are Data-Efficient
  Learners for Self-Supervised Video Pre-Training}.
\newblock \bibinfo{journal}{\emph{CoRR}}  \bibinfo{volume}{abs/2203.12602}
  (\bibinfo{year}{2022}).
\newblock
\urldef\tempurl%
\url{https://doi.org/10.48550/arXiv.2203.12602}
\showDOI{\tempurl}
\showeprint[arXiv]{2203.12602}


\bibitem[\protect\citeauthoryear{Tulyakov, Liu, Yang, and Kautz}{Tulyakov
  et~al\mbox{.}}{2018}]%
        {MocoGAN}
\bibfield{author}{\bibinfo{person}{Sergey Tulyakov}, \bibinfo{person}{Ming-Yu
  Liu}, \bibinfo{person}{Xiaodong Yang}, {and} \bibinfo{person}{Jan Kautz}.}
  \bibinfo{year}{2018}\natexlab{}.
\newblock \showarticletitle{MoCoGAN: Decomposing Motion and Content for Video
  Generation}. In \bibinfo{booktitle}{\emph{Proceedings of the IEEE Conference
  on Computer Vision and Pattern Recognition (CVPR)}}.
\newblock


\bibitem[\protect\citeauthoryear{Varol, Romero, Martin, Mahmood, Black, Laptev,
  and Schmid}{Varol et~al\mbox{.}}{2017}]%
        {varol17_surreal}
\bibfield{author}{\bibinfo{person}{G{\"u}l Varol}, \bibinfo{person}{Javier
  Romero}, \bibinfo{person}{Xavier Martin}, \bibinfo{person}{Naureen Mahmood},
  \bibinfo{person}{Michael~J. Black}, \bibinfo{person}{Ivan Laptev}, {and}
  \bibinfo{person}{Cordelia Schmid}.} \bibinfo{year}{2017}\natexlab{}.
\newblock \showarticletitle{Learning from Synthetic Humans}. In
  \bibinfo{booktitle}{\emph{CVPR}}.
\newblock


\bibitem[\protect\citeauthoryear{von K\"ugelgen, Sharma, Gresele, Brendel,
  Sch\"olkopf, Besserve, and Locatello}{von K\"ugelgen et~al\mbox{.}}{2021}]%
        {Kugelgen2021NeurIPS}
\bibfield{author}{\bibinfo{person}{Julius von K\"ugelgen},
  \bibinfo{person}{Yash Sharma}, \bibinfo{person}{Luigi Gresele},
  \bibinfo{person}{Wieland Brendel}, \bibinfo{person}{Bernhard Sch\"olkopf},
  \bibinfo{person}{Michel Besserve}, {and} \bibinfo{person}{Francesco
  Locatello}.} \bibinfo{year}{2021}\natexlab{}.
\newblock \showarticletitle{Self-Supervised Learning with Data Augmentations
  Provably Isolates Content from Style}. In \bibinfo{booktitle}{\emph{Advances
  in Neural Information Processing Systems}}, Vol.~\bibinfo{volume}{34}.
\newblock
\urldef\tempurl%
\url{https://proceedings.neurips.cc/paper/2021/hash/8929c70f8d710e412d38da624b21c3c8-Abstract.html}
\showURL{%
\tempurl}


\bibitem[\protect\citeauthoryear{Wang, Gao, Li, Lin, Ma, Cheng, Peng, Huang,
  Ji, and Sun}{Wang et~al\mbox{.}}{2021}]%
        {Wang_2021_CVPR}
\bibfield{author}{\bibinfo{person}{Jinpeng Wang}, \bibinfo{person}{Yuting Gao},
  \bibinfo{person}{Ke Li}, \bibinfo{person}{Yiqi Lin}, \bibinfo{person}{Andy~J.
  Ma}, \bibinfo{person}{Hao Cheng}, \bibinfo{person}{Pai Peng},
  \bibinfo{person}{Feiyue Huang}, \bibinfo{person}{Rongrong Ji}, {and}
  \bibinfo{person}{Xing Sun}.} \bibinfo{year}{2021}\natexlab{}.
\newblock \showarticletitle{Removing the Background by Adding the Background:
  Towards Background Robust Self-Supervised Video Representation Learning}. In
  \bibinfo{booktitle}{\emph{Proceedings of the IEEE/CVF Conference on Computer
  Vision and Pattern Recognition (CVPR)}}. \bibinfo{pages}{11804--11813}.
\newblock


\bibitem[\protect\citeauthoryear{Wang, Girshick, Gupta, and He}{Wang
  et~al\mbox{.}}{2018}]%
        {Nonlocal}
\bibfield{author}{\bibinfo{person}{Xiaolong Wang}, \bibinfo{person}{Ross
  Girshick}, \bibinfo{person}{Abhinav Gupta}, {and} \bibinfo{person}{Kaiming
  He}.} \bibinfo{year}{2018}\natexlab{}.
\newblock \showarticletitle{Non-Local Neural Networks}. In
  \bibinfo{booktitle}{\emph{Proceedings of the IEEE Conference on Computer
  Vision and Pattern Recognition (CVPR)}}.
\newblock


\bibitem[\protect\citeauthoryear{Wishart, Feunang, Marcu, Guo, Liang,
  V{\'a}zquez-Fresno, Sajed, Johnson, Li, Karu, et~al\mbox{.}}{Wishart
  et~al\mbox{.}}{2018}]%
        {wishart2018hmdb}
\bibfield{author}{\bibinfo{person}{David~S Wishart},
  \bibinfo{person}{Yannick~Djoumbou Feunang}, \bibinfo{person}{Ana Marcu},
  \bibinfo{person}{An~Chi Guo}, \bibinfo{person}{Kevin Liang},
  \bibinfo{person}{Rosa V{\'a}zquez-Fresno}, \bibinfo{person}{Tanvir Sajed},
  \bibinfo{person}{Daniel Johnson}, \bibinfo{person}{Carin Li},
  \bibinfo{person}{Naama Karu}, {et~al\mbox{.}}}
  \bibinfo{year}{2018}\natexlab{}.
\newblock \showarticletitle{HMDB 4.0: the human metabolome database for 2018}.
\newblock \bibinfo{journal}{\emph{Nucleic acids research}}
  \bibinfo{volume}{46}, \bibinfo{number}{D1} (\bibinfo{year}{2018}),
  \bibinfo{pages}{D608--D617}.
\newblock


\bibitem[\protect\citeauthoryear{Wu, Kirillov, Massa, Lo, and Girshick}{Wu
  et~al\mbox{.}}{2019}]%
        {wu2019detectron2}
\bibfield{author}{\bibinfo{person}{Yuxin Wu}, \bibinfo{person}{Alexander
  Kirillov}, \bibinfo{person}{Francisco Massa}, \bibinfo{person}{Wan-Yen Lo},
  {and} \bibinfo{person}{Ross Girshick}.} \bibinfo{year}{2019}\natexlab{}.
\newblock \bibinfo{title}{Detectron2}.
\newblock
  \bibinfo{howpublished}{\url{https://github.com/facebookresearch/detectron2}}.
\newblock


\bibitem[\protect\citeauthoryear{Xu, Yoon, Fuentes, and Park}{Xu
  et~al\mbox{.}}{2022}]%
        {DBLP:journals/corr/abs-2205-01491}
\bibfield{author}{\bibinfo{person}{Mingle Xu}, \bibinfo{person}{Sook Yoon},
  \bibinfo{person}{Alvaro Fuentes}, {and} \bibinfo{person}{Dong~Sun Park}.}
  \bibinfo{year}{2022}\natexlab{}.
\newblock \showarticletitle{A Comprehensive Survey of Image Augmentation
  Techniques for Deep Learning}.
\newblock \bibinfo{journal}{\emph{CoRR}}  \bibinfo{volume}{abs/2205.01491}
  (\bibinfo{year}{2022}).
\newblock
\urldef\tempurl%
\url{https://doi.org/10.48550/arXiv.2205.01491}
\showDOI{\tempurl}
\showeprint[arXiv]{2205.01491}


\bibitem[\protect\citeauthoryear{Yoo, Ahn, and Sohn}{Yoo et~al\mbox{.}}{2020}]%
        {CutBlur}
\bibfield{author}{\bibinfo{person}{Jaejun Yoo}, \bibinfo{person}{Namhyuk Ahn},
  {and} \bibinfo{person}{Kyung-Ah Sohn}.} \bibinfo{year}{2020}\natexlab{}.
\newblock \showarticletitle{Rethinking Data Augmentation for Image
  Super-resolution: A Comprehensive Analysis and a New Strategy}.
\newblock \bibinfo{journal}{\emph{arXiv preprint arXiv:2004.00448}}
  (\bibinfo{year}{2020}).
\newblock


\bibitem[\protect\citeauthoryear{Yun, Han, Oh, Chun, Choe, and Yoo}{Yun
  et~al\mbox{.}}{2019}]%
        {Yun_2019_ICCV}
\bibfield{author}{\bibinfo{person}{Sangdoo Yun}, \bibinfo{person}{Dongyoon
  Han}, \bibinfo{person}{Seong~Joon Oh}, \bibinfo{person}{Sanghyuk Chun},
  \bibinfo{person}{Junsuk Choe}, {and} \bibinfo{person}{Youngjoon Yoo}.}
  \bibinfo{year}{2019}\natexlab{}.
\newblock \showarticletitle{CutMix: Regularization Strategy to Train Strong
  Classifiers With Localizable Features}. In
  \bibinfo{booktitle}{\emph{Proceedings of the IEEE/CVF International
  Conference on Computer Vision (ICCV)}}.
\newblock


\bibitem[\protect\citeauthoryear{Yun, Oh, Heo, Han, and Kim}{Yun
  et~al\mbox{.}}{2020}]%
        {VideoMix}
\bibfield{author}{\bibinfo{person}{Sangdoo Yun}, \bibinfo{person}{Seong~Joon
  Oh}, \bibinfo{person}{Byeongho Heo}, \bibinfo{person}{Dongyoon Han}, {and}
  \bibinfo{person}{Jinhyung Kim}.} \bibinfo{year}{2020}\natexlab{}.
\newblock \showarticletitle{VideoMix: Rethinking Data Augmentation for Video
  Classification}.
\newblock \bibinfo{journal}{\emph{CoRR}}  \bibinfo{volume}{abs/2012.03457}
  (\bibinfo{year}{2020}).
\newblock
\showeprint[arxiv]{2012.03457}
\urldef\tempurl%
\url{https://arxiv.org/abs/2012.03457}
\showURL{%
\tempurl}


\bibitem[\protect\citeauthoryear{Zhang, Ciss{\'{e}}, Dauphin, and
  Lopez{-}Paz}{Zhang et~al\mbox{.}}{2018}]%
        {Zhang_iclr18}
\bibfield{author}{\bibinfo{person}{Hongyi Zhang}, \bibinfo{person}{Moustapha
  Ciss{\'{e}}}, \bibinfo{person}{Yann~N. Dauphin}, {and} \bibinfo{person}{David
  Lopez{-}Paz}.} \bibinfo{year}{2018}\natexlab{}.
\newblock \showarticletitle{mixup: Beyond Empirical Risk Minimization}. In
  \bibinfo{booktitle}{\emph{6th International Conference on Learning
  Representations, {ICLR} 2018, Vancouver, BC, Canada, April 30 - May 3, 2018,
  Conference Track Proceedings}}. \bibinfo{publisher}{OpenReview.net}.
\newblock
\urldef\tempurl%
\url{https://openreview.net/forum?id=r1Ddp1-Rb}
\showURL{%
\tempurl}


\end{thebibliography}

\end{document}